\documentclass[letterpaper]{article} % DO NOT CHANGE THIS
\usepackage[]{aaai2026}  % DO NOT CHANGE THIS
\usepackage{times}  % DO NOT CHANGE THIS
\usepackage{helvet}  % DO NOT CHANGE THIS
\usepackage{courier}  % DO NOT CHANGE THIS
\usepackage[hyphens]{url}  % DO NOT CHANGE THIS
\usepackage{graphicx} % DO NOT CHANGE THIS
\usepackage{amssymb}
\usepackage{amsmath}
\usepackage{multirow}
\usepackage{booktabs} 

\usepackage[table]{xcolor}
\usepackage{arydshln} 
\usepackage{pgfplots}

\usepackage{natbib}  % DO NOT CHANGE THIS AND DO NOT ADD ANY OPTIONS TO IT
\usepackage{caption} % DO NOT CHANGE THIS AND DO NOT ADD ANY OPTIONS TO IT
\usepackage{algorithm} % http://ctan.org/pkg/algorithms
\usepackage{algpseudocode} % 

\usepackage{tikz}
\usetikzlibrary{shapes.geometric, arrows, positioning, fit, backgrounds, matrix, arrows.meta, shadows}
\usetikzlibrary{decorations.pathmorphing}
\usepackage{soul}
\usetikzlibrary{pgfplots.groupplots}
\usetikzlibrary{patterns} 
\usepackage{listings}
\usepackage{tcolorbox}
\usepackage{mathtools}
\usepackage{subcaption}

\usepackage{newfloat}
\usepackage{listings}
\DeclareCaptionStyle{ruled}{labelfont=normalfont,labelsep=colon,strut=off} % DO NOT CHANGE THIS
\floatstyle{ruled}
\newfloat{listing}{tb}{lst}{}
\floatname{listing}{Listing}

\usepackage{hyperref} 

\title{GRAM-R$^2$: Self-Training Generative Foundation Reward Models \\ for Reward Reasoning}

\author{
    Chenglong Wang\textsuperscript{\rm 1}\footnote{Authors contributed equally.} \quad 
    Yongyu Mu\textsuperscript{\rm 1}$^*$\footnote{Work was done when Yongyu Mu was interning at Pattern Recognition Center, WeChat AI, Tencent Inc.}
    \quad 
    Hang Zhou\textsuperscript{\rm 1}
    \quad 
    Yifu Huo\textsuperscript{\rm 1}
    \quad 
    Ziming Zhu\textsuperscript{\rm 1} \\ 
    Jiali Zeng\textsuperscript{\rm 2}
    \quad 
    Murun Yang\textsuperscript{\rm 1}
    \quad 
    Bei Li\textsuperscript{\rm 1}
    \quad 
    Xiaoyang Hao\textsuperscript{\rm 1} \\
    Chunliang Zhang\textsuperscript{\rm 1}
    \quad 
    Fandong Meng\textsuperscript{\rm 2}
    \quad 
    Jingbo Zhu\textsuperscript{\rm 1}
    \quad 
    Tong Xiao\textsuperscript{\rm 1}\footnote{Corresponding author.} 
}
\affiliations{
    \textsuperscript{\rm 1} School of Computer Science and Engineering, Northeastern University, Shenyang, China \\
    \textsuperscript{\rm 2} Pattern Recognition Center, WeChat AI, Tencent Inc., China \\
     \texttt{clwang1119@gmail.com \quad xiaotong@mail.neu.edu.cn}
\begin{center}
\raisebox{-0.3ex}{\includegraphics[height=1.1em]{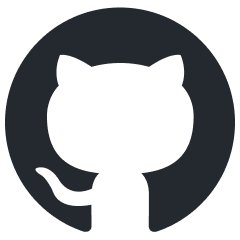}}
\textcolor{blue!70}{\href{https://github.com/NiuTrans/GRAM/tree/main/extensions/GRAM-RR}{Code}} \quad \quad 
\raisebox{-0.9ex}{\includegraphics[height=1.45em]{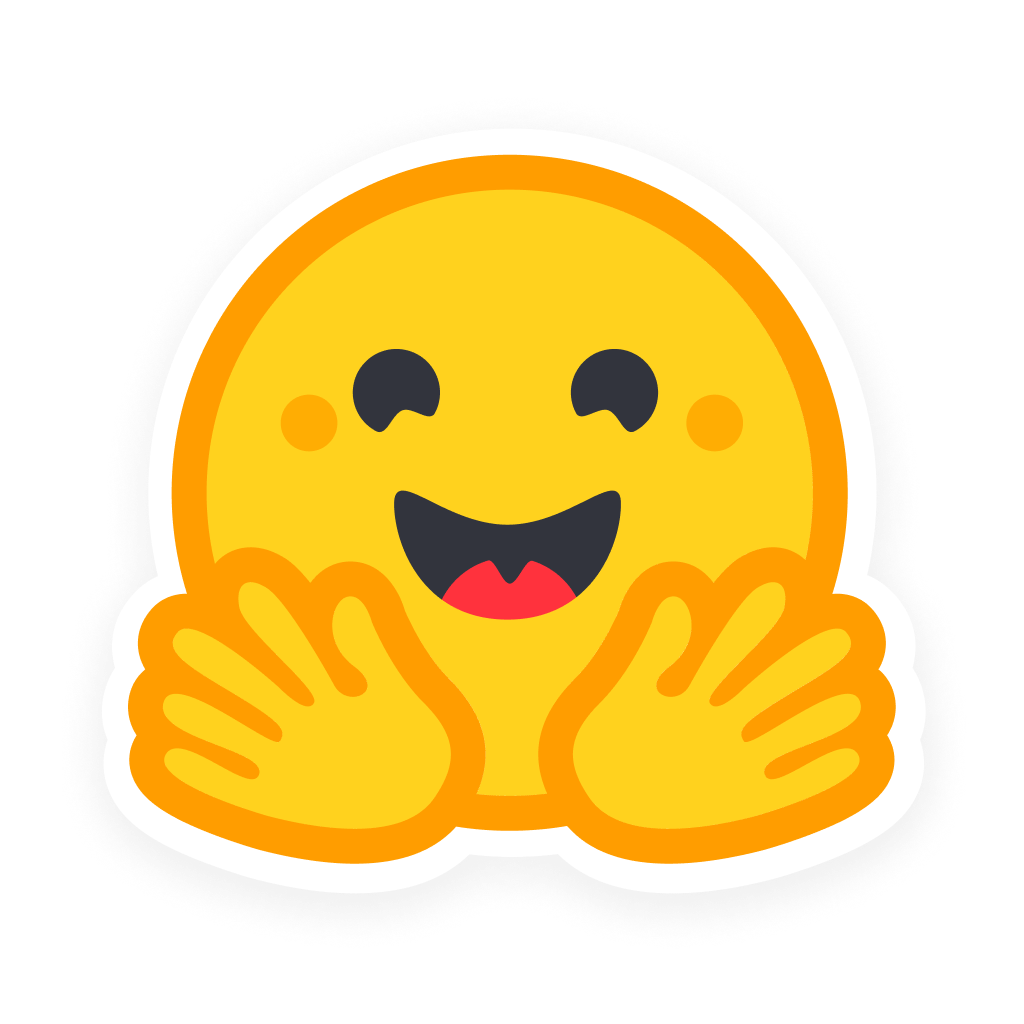}}
\textcolor{blue!70}{\href{https://huggingface.co/collections/wangclnlp/gram-rr-68b2c8d83e35625bc57c7e81}{Datasets\&Models}}
\end{center}
\vspace{-4mm}
}

\usepackage{bibentry}
\begin{document}

\maketitle

\begin{abstract}
Significant progress in reward modeling over recent years has been driven by a paradigm shift from task-specific designs towards generalist reward models. Despite this trend, developing effective reward models remains a fundamental challenge: the heavy reliance on large-scale labeled preference data. Pre-training on abundant unlabeled data offers a promising direction, but existing approaches fall short of instilling explicit reasoning into reward models. To bridge this gap, we propose a self-training approach that leverages unlabeled data to elicit reward reasoning in reward models. Based on this approach, we develop GRAM-R$^2$, a generative reward model trained to produce not only preference labels but also accompanying reward rationales. GRAM-R$^2$ can serve as a foundation model for reward reasoning and can be applied to a wide range of tasks with minimal or no additional fine-tuning. It can support downstream applications such as response ranking and task-specific reward tuning. Experiments on response ranking, task adaptation, and reinforcement learning from human feedback demonstrate that GRAM-R$^2$ consistently delivers strong performance, outperforming several strong discriminative and generative baselines.

\end{abstract}

\section{Introduction}
% Reward models are central to reinforcement learning, as they define the objectives that agents aim to optimize \cite{sutton1998reinforcement}. 
Reward models are a cornerstone of aligning large language models (LLMs) with human preferences during post-training. Typically, a reward model is trained to encode these preferences, and the LLM is subsequently fine-tuned to maximize the reward signal it provides. This paradigm is first exemplified by reinforcement learning from human feedback (RLHF) \cite{stiennon2020learning}. More recently, the use of reward models has expanded beyond training into inference, where they are used to re-rank candidate responses. This approach has emerged as a strategy in studies on inference-time scaling laws \cite{wu2024empirical,li2025system}.

The dominant approach to developing reward models is to collect a dataset of training examples demonstrating correct behavior for desired human preferences in a specific task, train a model to imitate these behaviors, and then test its performance to align LLMs with independent and identically distributed examples. While this approach has proven successful for aligning LLMs in narrow contexts \cite{stiennon2020learning,xu2024contrastive}, its application is limited to these tasks. 
% such as summarization \cite{stiennon2020learning} and machine translation \cite{xu2024contrastive}, its application is limited to these tasks. 
As the field progresses towards artificial general intelligence (AGI), a paradigm shift is necessary: moving towards generalist reward models that can generalize across a wide range of tasks to facilitate the broader alignment of AI systems with human preferences.

Labeling multi-task, large-scale preference data offers a strategy to enhance generalist performance \cite{cui2023ultrafeedback,wang2023helpsteer,wang2024helpsteer}. However, from a multi-task learning perspective, each labeled example is drawn from a task-specific distribution, and current reward models typically require hundreds or thousands of labeled examples to learn functions that generalize well across tasks \cite{zhang2021survey}. This reliance on labeled data poses a significant bottleneck, making it challenging to scale reward model training to the level of LLM training.
% This motivates exploring additional setups for training generalist reward models.

A promising direction is to pre-train on unlabeled data before fine-tuning on a smaller labeled set. This two-stage paradigm first equips the model with implicit knowledge of human preferences from unlabeled data, such as input-response pairs, and then fine-tunes it using labeled data. Since the pre-training stage does not depend on large-scale labeled datasets, it is highly scalable. Under this paradigm, foundation reward models such as GRAM \cite{wang2025gram} and POLAR \cite{dou2025pre} have emerged. However, while these foundation models effectively learn \textit{what} humans prefer, they do not capture the explicit reasoning behind \textit{why} those preferences are held during the pre-training process. This limitation prevents them from leveraging the strong reasoning capabilities inherent to the LLM backbone. More importantly, another line of work has demonstrated that incorporating explicit reasoning (referred to as \textit{reward reasoning}) into reward models can substantially improve model performance \cite{chen2025rm,guo2025reward}.

In this paper, we connect these two lines of work and extend the pre-training stage to incorporate reward reasoning explicitly. Our goal is to endow foundation reward models with the capability to perform reward reasoning across a wide range of downstream tasks, either without fine-tuning or with only minimal task-specific supervision. 
To train this model, we propose a self-training approach designed to elicit reward reasoning using labeled data that lacks rationales (referred to as \textit{rationale-free labeled data}) and vast amounts of unlabeled data. This approach can circumvent the need for expensive rationale-based annotations, thus ensuring the scalability required for building foundation reward models.
Specifically, we first train a preference-proving model conditioned on an input, a response pair, and a preference label, which generates a proof explaining why the labeled preference holds. For rationale-free labeled data, we use this preference-proving model to synthesize rationales for each example. For unlabeled data, we allow the reward model to enhance its reward reasoning capability through a self-training loop iteratively: 1) the reward model predicts preference labels for unlabeled data; 2) the preference-proving model generates corresponding rationales; and 3) the reward model is updated using the synthesized data. Notably, our self-training process allows the reward model to scale up its reward reasoning by leveraging vast unlabeled data. 

% To achieve this, we develop a generative reward model that, given an input and a pair of responses, not only predicts the superior response but also provides an explicit rationale for it. We propose a self-training approach for training this model. This approach harnesses the potential of vast, open datasets, including both labeled data that lacks rationales (referred to as \textit{rationale-free labeled data}) and unlabeled data, to elicit reward reasoning in reward models

We introduce the resulting model as a \textbf{\underline{G}}enerative foundation \textbf{\underline{R}}ew\textbf{\underline{A}}rd \textbf{\underline{M}}odel for \textbf{\underline{R}}eward \textbf{\underline{R}}easoning  (GRAM-R$^2$). It can be directly applied to downstream tasks such as response ranking or further fine-tuned with a small amount of task-specific data. In our experiments, we evaluate GRAM-R$^2$ under three settings: response ranking, task adaptation, and RLHF. Across all test cases, GRAM-R$^2$ consistently exhibits a strong reward reasoning capability with little or no additional fine-tuning, and significantly outperforms both discriminative and generative baselines. For instance, when using LLaMA-3.1-8B-Instruct as the backbone, GRAM-R$^2$ achieves gains of 10.1 and 6.9 points in average accuracy on RM-Bench over vanilla discriminative and generative reward models, respectively. These results demonstrate that strong reasoning capabilities can be elicited from rationale-free labeled and unlabeled data.

\section{Related Work}
In recent years, reward models have played a critical role in aligning LLMs with human preferences \cite{stiennon2020learning,huo2025heal}. Pre-training reward models on unlabeled data has proven effective for improving performance \cite{wang2025gram,dou2025pre}. However, in this process, they never focus on cultivating the reward model’s ability to perform reward reasoning.

\paragraph{Reward Modeling.}
Reward models, typically trained on human preference data, are central to RLHF and other alignment strategies like DPO and rejection sampling \cite{lee2021discriminative,rafailov2023direct,chu2023qwen,wang2024improving,zhou2024prior,wang2025error}. 
% More recently, researchers have extended the use of reward models beyond training and into inference \cite{wu2024empirical,li2025system}. 
Recent works on improving reward models could be classified into three groups. The first group focused on large-scale, high-quality training data, developing either task-specific datasets \cite{stiennon2020learning,xu2024contrastive} or more general preference datasets \cite{cui2023ultrafeedback}. The second group explored stronger reward modeling approaches \cite{coste2023reward,min2024dynamic}. Notably, researchers have shown that integrating explicit reasoning into reward models is crucial for improving alignment performance \cite{chen2025rm,guo2025reward}. Although reward modeling through these approaches effectively captures human preferences, they often rely heavily on complex reinforcement learning and labeled data. To alleviate this, a third line of work has emerged that leverages unlabeled data to pre-train foundation reward models, such as GRAM \cite{wang2025gram} and POLAR \cite{dou2025pre}. However, these approaches overlook the development of reward reasoning capabilities, thereby limiting the model to exploit the reasoning potential of the LLM backbone fully. This motivates us to train a foundation reward model with unlabeled data for reward reasoning.

\paragraph{Self-Training.}
Self-training \cite{scudder1965probability,han2019deep,xie2020self,wang2021progressive} is a classic semi-supervised learning framework. The basic idea is to employ model predictions on unlabeled data to generate pseudo-labels. These pseudo-labeled examples are then used to augment the original training set, enabling the model to improve its performance by leveraging large-scale unlabeled corpora without requiring additional human annotation. Such a guiding principle has shown
empirical success in diverse domains such as computer vision \cite{yalniz2019billion,zoph2020rethinking}, natural language processing \cite{yeo2024self,zhang2024rest,luo2025self}, and lifelong learning \cite{lee2019overcoming}. Here, we extend this idea to training reward models and show that self-training with large-scale unlabeled data can effectively scale up reward reasoning in reward models. To our knowledge, this is the first work to apply self-training to reward model training.

\section{Preliminaries}
% We begin by reviewing the foundations of reward model training and then describe common applications of trained reward models.

\subsection{Reward Model Training}
In LLMs literature, a reward model is typically written as a function $r_{\phi}(x,y)$, where $\phi$ is the set of model parameters, $x$ is the input, and $y$ is the response. Throughout this work, an \textit{input} can be an arbitrary token sequence fed into an LLM, such as ``\textit{What is the capital of France?}'', and a \textit{response} is the token sequence produced by the LLM as a result of that input, such as ``\textit{Paris}''. To date, mainstream reward model architectures can be broadly categorized into two types: \textit{discriminative} and \textit{generative}.

\paragraph{Discriminative Reward Models.} 
Discriminative reward models compute scores directly as scalar outputs from a classification architecture. Such a model typically consists of a Transformer decoder without a Softmax layer. The concatenated input–response $[x, y]$ is passed via a pre-trained LLM, and the final-layer hidden representations are used to compute a scalar score. This model can be trained through a Bradley-Terry loss function \cite{bradley1952rank}:
\begin{eqnarray}
\mathcal{L}_{\mathrm{d}} & = & - \mathbb{E}_{\tiny (x,y_{a},y_{b})\sim D_{r}} \nonumber \\ 
& & \big[ \log (\sigma (r_{\phi}(x,y_{a})-r_{\phi}(x,y_{b}))) \big]
\end{eqnarray}
where $D_{r}$ is the training dataset consisting of tuples of input $x$ and response pair $(y_{a},y_{b})$ with the preference $y_{a} \succ y_{b}$. While this loss function considers pairwise ranking between responses, the trained reward model is used as a scoring function that assigns a numerical reward $r_{\phi}(x,y)$ to each response $y$, along with its corresponding input $x$.

\paragraph{Generative Reward Models.} 
While discriminative reward models have demonstrated success, this scoring approach fails to fully leverage the text generation capabilities that LLMs are fundamentally designed for \cite{zhang2024generative}. To address this limitation, recent studies have increasingly focused on developing generative reward models \cite{liang2025generative}.
These models produce reward signals via natural language generation. 
% As shown in Figure~\ref{fig:reward_model_arch}, 
Specifically, they use an LLM to generate preference-related tokens, given a natural language prompt $c$ and a tuple $(x, y_a, y_b)$. The prompt describes the task in natural language, and the model predicts a label token $w$ that aligns with the human preference $l$, where $l = \text{A}$ denotes preference for $y_a$, and $l = \text{B}$ indicates preference for $y_b$. The model can be trained by
\begin{eqnarray}
\mathcal{L}_{\mathrm{g}} &= & - \mathbb{E}_{(c,x,y_a,y_b,l) \sim D_r}
\big[\log \pi_{\phi}(w=l|s)\big] \label{eq:gem-reward-modeling}
\end{eqnarray}
where $s$ denotes the string $[c,x,y_a,y_b]$, and $\pi_{\phi}(\cdot)$ denotes the probability of token prediction by the LLM.

\begin{figure}[!t]
    \centering
    \pgfdeclarepatternformonly{soft crosshatch}{\pgfqpoint{-1pt}{-1pt}}{\pgfqpoint{6pt}{6pt}}{\pgfqpoint{5pt}{5pt}}%
{
  \pgfsetstrokeopacity{0.3}
  \pgfsetlinewidth{0.4pt}
  \pgfpathmoveto{\pgfqpoint{4.5pt}{0pt}}
  \pgfpathlineto{\pgfqpoint{0pt}{4.5pt}}
  \pgfpathmoveto{\pgfqpoint{0pt}{0pt}}
  \pgfpathlineto{\pgfqpoint{4.5pt}{4.5pt}}
  \pgfusepath{stroke}
}

\begin{tikzpicture}

\def\ssep{1cm}
\def\nodewidth{3*\ssep}
\def\seg{0.3cm}

\begin{scope}[yshift=-4.0cm]

\node [anchor=south west,text width=2.3*\nodewidth,minimum height=1*\ssep,draw,thick,align=center] (llm) at (0,0) {\normalsize{Transformer Decoder (LLM)}};

\node [anchor=north west,minimum width=0.35*\nodewidth,,minimum height=0.45cm,rounded corners=2pt,draw,fill=gray!10] (input) at ([yshift=-\seg]llm.south west) {\scriptsize{$c$}};
\node [anchor=west,minimum width=0.35*\nodewidth,,minimum height=0.45cm,rounded corners=2pt,draw,fill=gray!10] (input2) at ([xshift=3pt]input.east) {\scriptsize{$x$}};
\node [anchor=west,minimum width=0.35*\nodewidth,,minimum height=0.45cm,rounded corners=2pt,draw,fill=orange!20] (input3) at ([xshift=3pt]input2.east) {\scriptsize{$y_{a}$}};
\node [anchor=west,minimum width=0.35*\nodewidth,,minimum height=0.45cm,rounded corners=2pt,draw,fill=orange!20] (input4) at ([xshift=3pt]input3.east) {\scriptsize{$y_{b}$}};
\draw [->] ([yshift=1pt]input.north) -- ([yshift=\seg-1pt]input.north);
\draw [->] ([yshift=1pt]input2.north) -- ([yshift=\seg-1pt]input2.north);
\draw [->] ([yshift=1pt]input3.north) -- ([yshift=\seg-1pt]input3.north);
\draw [->] ([yshift=1pt]input4.north) -- ([yshift=\seg-1pt]input4.north);

\node [anchor=center] (inputlabel) at ([yshift=-0.20cm]input.south) {\scriptsize{Prompt}};
\node [anchor=center] (input2label) at ([yshift=-0.20cm]input2.south) {\scriptsize{Input}};
\node [anchor=center,scale=0.9] (input3label) at ([yshift=-0.20cm]input3.south) {\scriptsize{Response A}};
\node [anchor=center, scale=0.9] (input4label) at ([yshift=-0.20cm]input4.south) {\scriptsize{Response B}};

\node [anchor=south east,minimum width=4cm,minimum height=0.6cm,rounded corners=0pt,draw,fill=blue!5,align=left,scale=0.85] (output1) at ([yshift=\seg,xshift=0*\seg]llm.north east) {
\scriptsize $<$\textit{think}$>$ \\[-4pt]
\scriptsize \textbf{Feedback}: Response A... . Response B... . \\[-4pt]
\scriptsize \textbf{Comparison}: Response B is better, because... \\[-4pt]
\scriptsize \textbf{Conclusion}: Response B is better. \\[-4pt]
\scriptsize $<$\textit{/think}$>$ \\[-2pt] 
\scriptsize $<$\textit{answer}$>$ \textbf{B} $<$\textit{/answer}$>$
};
% \node [anchor=south east,minimum width=1cm,minimum height=0.45cm,rounded corners=0pt, pattern = soft crosshatch] (output1) at ([yshift=\seg,xshift=-3*\seg]llm.north east) {};

% \node [anchor=center] (output1label) at ([yshift=0.17cm]output1.north) {\tiny{rationale}};

% \node [anchor=south east,minimum width=0.45cm,minimum height=0.45cm,rounded corners=0pt,draw,fill=blue!10] (output2) at ([yshift=\seg,xshift=-0.8*\seg]llm.north east) {\scriptsize{$l$}};

\draw[decorate,decoration={brace,amplitude=3pt}] (3,1.75) -- (3,2.85) node[midway,xshift=-0.8cm] {\scriptsize Rationale $z$};

\draw[decorate,decoration={brace,amplitude=1pt}] (3,1.35) -- (3,1.65) node[midway,xshift=-1.1cm] {\scriptsize Preference Label $l$};

\draw [->] ([yshift=-\seg+1pt]output1.south) -- ([yshift=-1pt]output1.south);
% \draw [->] ([yshift=-\seg+1pt]output2.south) -- ([yshift=-1pt]output2.south);

% \node [anchor=east] (tokenlabel) at (output.west) {\tiny{next token (`A' or `B')}};

% \node [anchor=south west,align=left] (loss) at ([yshift=0.9cm]llm.north west) {\scriptsize{Minimizing the negative probability of token prediction:}\\ \scriptsize{$-\log \pi_{\phi}(w=\text{A}|[c,x,y_a,y_b])$}};

% \node [anchor=north] (caption) at ([xshift=0.8cm,yshift=-0.4cm]input2.south east) {\small{(b) Generative Models (Trained as LLMs)}};

\end{scope}
\end{tikzpicture}
    \vspace{-2mm}
    \caption{
    Architecture of the Generative Reward Model.
    The generative reward model utilizes a pre-trained LLM to predict a preference label from a given prompt directly. Optionally, it can incorporate reward reasoning before generating the final preference label prediction.
    }
    \vspace{-0.4cm}
    \label{fig:reward_model_arch}
\end{figure}
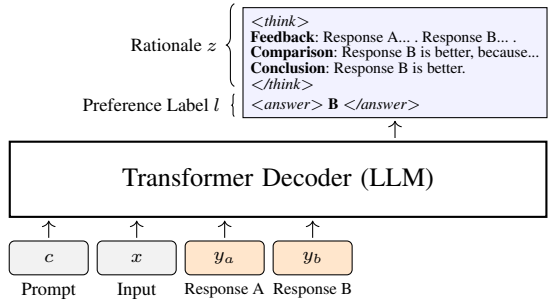

Recent studies have shown that framing reward prediction as a reasoning task can further leverage the powerful reasoning capabilities of LLMs to improve reward modeling performance \cite{chen2025rm,guo2025reward}. In these works, the model is trained to generate explicit reward reasoning (\textit{e.g.}, analyzing and evaluating the responses individually) before predicting the final preference label as shown in Figure~\ref{fig:reward_model_arch}. Let $z$ denote this rationale, a natural language justification for the preference label. The model first generates $z$ conditioned on the input string $s$, and then predicts the preference label based on both the context and the generated rationale. 
In this process, it can be trained to generate both the rationale and the final label via the following objective:
\begin{eqnarray}
\mathcal{L}_{\mathrm{g}} &=& - \mathbb{E}_{(c, x, y_a, y_b, l, z) \sim D_p} \nonumber \\
&& \big[ \log \pi_{\phi}(z | s) + \log \pi_{\phi}(w = l | s, z) \big]
\end{eqnarray}
where $\mathcal{D}_p$ is a set of annotated data containing both a preference label $l$ and a corresponding labeled rationale $z$. Note that although incorporating reward reasoning significantly improves the performance of reward models, it presents a non-trivial challenge: it requires costly human annotations that include not only preference labels but also their corresponding detailed rationales. 

\subsection{Applying Reward Models}
Three applications of foundation reward models can be considered in LLMs.  
A straightforward application is response ranking, where several responses are given, and we score and rank these responses. This approach is widely used in reranking settings, such as best-of-$n$ sampling, where the highest-scoring response among $n$ candidates is selected based on reward scores \cite{lee2021discriminative,fernandes2022quality,gao2023scaling}.

A second application of reward models is to provide learning signals for fine-tuning LLMs toward human preferences in RLHF, typically through algorithms such as Proximal Policy Optimization (PPO) \cite{ouyang2022training, wang2023improved}.

A third application is that when task-specific human preference data is available, the reward model can be further fine-tuned to better align with that particular task \cite{wang2025worldpm,dou2025pre}. The adapted reward model can then be used in downstream applications such as RLHF or response ranking.

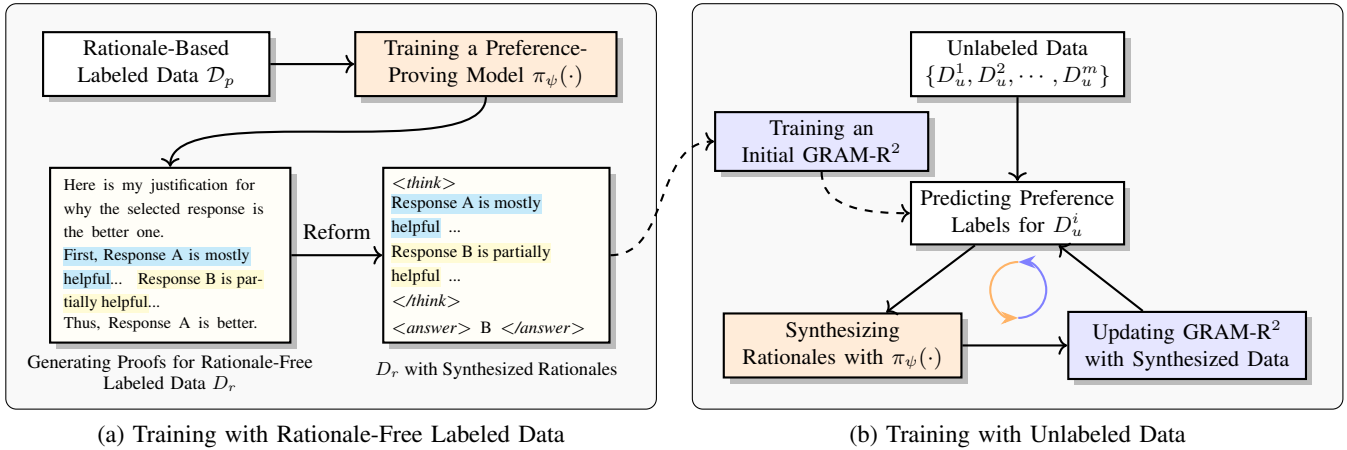
\begin{figure*}[!t]
    \centering
    \begin{tikzpicture}[scale=0.93, transform shape]
    % 样式定义
    \tikzset{
        block/.style={rectangle, draw, thick, rounded corners, text width=2em, text centered, minimum height=3em, inner sep=2.5pt},
        rblock/.style={rectangle, draw, thick, text centered},
        dashed_block/.style={rectangle, draw, thick, dashed, rounded corners},
        arrow/.style={-To, thick}, % 使用 arrows.meta 的 Stealth
        circ_arrow/.style={-Stealth, thick, rounded corners},
        background_L/.style={rectangle, rounded corners, draw=black},
        background_U/.style={rectangle, rounded corners, draw=black}
    }

    % =============================================
    % 1. 先定义两个固定大小、并排的背景框
    % =============================================
    \node[background_L, minimum width=9.3cm, minimum height=5.8cm, fill=gray!5] (background1) at (0,0) {};
    \node[background_U, minimum width=9.3cm, minimum height=5.8cm, fill=gray!5, right=0.5cm of background1] (background2) {};

    % =============================================
    % 2. 左侧内容 (Training With Labeled Data)
    % =============================================
    % 标题 (已移动到下方)
    \node[text width=8cm, align=center, anchor=north, at=(background1.south), yshift=-0.1cm] (title_L) {(a) Training with Rationale-Free Labeled Data};

    % 使用 Matrix 创建数据块网格
    \node[
        rblock,
        below=0.4cm of background1.north,
        text width=3.0cm,
        xshift=-2.5cm,
        fill=white,        % Adds a light gray background color
        minimum height=0.9cm,
        drop shadow
    ] (data_box) {
        \footnotesize
        Rationale-Based Labeled Data $\mathcal{D}_{p}$
    };    

    % \node[rblock, right=1.2cm of data_box, text width=4cm, fill=white, drop shadow, scale=0.9] (train_model) {
    % % \setlength{\baselineskip}{2pt}
    % \footnotesize Training Preference-Proving Model to generate proof
    % };
        
    \node[rblock, right=1.2cm of data_box, text width=3.5cm, fill=orange!15, drop shadow] (train_model) {
    \footnotesize Training a Preference-Proving Model $\pi_\psi(\cdot)$
    };

    % --- 步骤1: 创建一个空的 "proof" 容器框 ---
    \node[rblock, below=1cm of data_box, text width=3.2cm, minimum height=2.5cm, fill=yellow!3, drop shadow, xshift=0.2cm] (proof) {};
    
    % --- 步骤2: 在容器内部放置标题、横线和正文 ---
    
    % 放置标题 "Proof"
    \node[font=\footnotesize, scale=0.85, text width=5cm, align=center] (proof_title) at (proof.south) [yshift=-14pt] {Generating Proofs for Rationale-Free \\ Labeled Data $D_{r}$};
    
    % 绘制横线 (在标题下方)
    % \draw ([yshift=-2pt]proof_title.south west) -- ([yshift=-2pt]proof_title.south east);
    
    % 放置正文 (在容器的正中心)
    \node[align=left, text width=3.2cm, scale=0.95] at (proof.center) {
        \scriptsize
        Here is my justification for why the selected response is the better one. \\
        \sethlcolor{cyan!20}
        \hl{First, Response A is mostly helpful}...
        \sethlcolor{yellow!20}
        \hl{Response B is partially helpful}... \\[-2pt]
        Thus, Response A is better.
    };
        
    % --- 对 "rationale" 节点应用完全相同的逻辑 ---
    
    % 创建一个空的 "rationale" 容器框
    \node[rblock, right=1.3cm of proof, text width=3cm, minimum height=2.5cm, fill=yellow!3, drop shadow] (rationale) {};
    
    % 放置标题 "Rationale"
    \node[font=\footnotesize, scale=0.85] (rationale_title) at (rationale.south) [yshift=-13pt] {$D_{r}$ with Synthesized Rationales};
    
    % 绘制横线
    % \draw ([yshift=-2pt]rationale_title.south west) -- ([yshift=-2pt]rationale_title.south east);
    
    % 放置正文
    \sethlcolor{cyan!20}
    \node[align=left, text width=3.0cm] at (rationale.center) {
        \scriptsize
        $<$\textit{think}$>$ \\[-2pt] 
        \hl{Response A is mostly} \\ 
        \hl{helpful} ... \\
        \sethlcolor{yellow!20}
        \hl{Response B is partially} \\ 
        \hl{helpful} ... \\
         $<$\textit{/think}$>$ \\
         $<$\textit{answer}$>$ B $<$\textit{/answer}$>$
    };
    
    % =============================================
    % 3. 右侧内容 (Training With Unlabeled Data)
    % =============================================
    % 标题 (已移动到下方)
    \node[text width=8cm, align=center, anchor=north, at=(background2.south), yshift=-0.1cm] (title_U) {(b) Training with Unlabeled Data};

    % 右侧的节点布局
    \node[rblock, below=0.4cm of background2.north, text width=8em, fill=white, minimum height=0.7cm, drop shadow] (unlabeled_data) {
    \footnotesize Unlabeled Data $\{D^1_{u},D^2_{u}, \cdots, D^m_{u}\}$};
    \node[rblock, below=1.2cm of unlabeled_data, text width=8em, fill=white, minimum height=0.7cm, drop shadow] (predict) {\footnotesize Predicting Preference Labels for $D^{i}_{u}$};
    \node[rblock, below=0.20cm of unlabeled_data, text width=8em, fill=blue!10, xshift=-2.8cm, minimum height=0.7cm, drop shadow] (initialize) {
    \footnotesize Training an Initial GRAM-R$^2$
    };
    \node[rblock, below=0.96cm of predict, text width=9em, fill=orange!15, xshift=-2.5cm, minimum height=0.7cm, drop shadow] (generate) {\footnotesize Synthesizing \\ \footnotesize Rationales with $\pi_{\psi}(\cdot)$};
    \node[rblock, right=1.5cm of generate, text width=9em, fill=blue!10, minimum height=0.7cm, drop shadow] (update) {\footnotesize Updating GRAM-R$^2$ with Synthesized Data};
    
    % =============================================
    % 4. 绘制所有箭头
    % =============================================
    % 左侧箭头
    \draw[arrow] (data_box) -- (train_model);
    \draw[arrow] (train_model.south) to[out=-90, in=90, , out distance=1cm, in distance=1cm] (proof.north);
    \draw[arrow] (proof) -- (rationale)
    node [above=0.1cm, midway, font=\footnotesize] {Reform};

    % 右侧箭头
    \draw[arrow, dashed] (initialize.south) to[out=-90, in=180] (predict.west);
    \draw[arrow] (unlabeled_data) -- (predict);
    \draw[arrow] (predict) -- (generate);
    \draw[arrow] (generate) -- (update);
    \draw[arrow] (update) -- (predict);
    % \node[scale=2.5] at ([xshift=-0.7cm, yshift=0.30cm]update.north west) {$\circlearrowleft$};
    % --- 替换原来的 \node 命令 ---

    % 1. 在目标位置定义一个中心点 (C)
    \coordinate (C) at ([xshift=-0.7cm, yshift=0.30cm]update.north west);
    
    % 2. 绘制上半圆的箭头 (例如，从右到左，蓝色)
    \draw[orange!60, -Stealth, thick] (C) ++(90:0.4cm) arc(90:270:0.4cm);
    
    % 3. 绘制下半圆的箭头 (例如，从左到右，红色)
    \draw[blue!50, -Stealth, thick] (C) ++(-90:0.4cm) arc(-90:90:0.4cm);
        
    % 连接左右的箭头
    \draw[arrow, dashed] (rationale.east) to[out=0, in=180] (initialize.west);

\end{tikzpicture}
    \vspace{-0.7cm}
    \caption{
    An overview of the self-training approach for GRAM-R$^2$. The process begins by training a preference-proving model on a small, rationale-based seed dataset of approximately 40.5K examples. This model is then used to synthesize rationales for a larger, rationale-free labeled dataset of 1M examples, which in turn is used to train the initial GRAM-R$^2$ model. Subsequently, GRAM-R$^2$ undergoes three iterations of self-training, using a new batch of 0.5M unlabeled examples in each iteration.}
    \vspace{-0.2cm}
    \label{fig:main_figure}
\end{figure*}

\section{Our Method}
In this section, we present a \underline{\textbf{G}}enerative foundation \underline{\textbf{R}}ew\underline{\textbf{A}}rd \underline{\textbf{M}}odel for \underline{\textbf{R}}eward \underline{\textbf{R}}easoning (GRAM-R$^2$). An overview of the GRAM-R$^2$ training process is shown in Figure~\ref{fig:main_figure}. As illustrated, we first train a preference-proving model and then utilize it to perform iterative self-training to pre-train GRAM-R$^2$, enabling it to scale up its reward reasoning using vast rationale-free labeled data and unlabeled data.

\subsection{Preference-Proving Model Training}
While a considerable amount of labeled preference data exists, it often lacks the very rationales needed to train generative reward models in the art of reward reasoning. To unlock the full potential of this data, we propose a preference-proving model that can automatically generate textual proofs for the provided preference labels.

\paragraph{Task Definition.} 
Given an example $(s, l)$ from a rationale-free labeled dataset $\mathcal{D}_r$, the objective of the preference-proving model is to generate a textual proof $\hat{z}$ that justifies the preference label $l$. We define the preference-proving model as a conditional LLM:
\begin{eqnarray}
\pi_\psi: (s, l) \mapsto \hat{z}
\end{eqnarray}
where $\psi$ denotes the model parameters. To train the model, we minimize the negative log-likelihood of generating the ground-truth rationale:
\begin{eqnarray}
    \mathcal{L}_{\text{p}} &=& - \mathbb{E}_{(c, x, y_a, y_b, l, z) \sim \mathcal{D}_{p}} \left[ \log \pi_\psi(\hat{z} \mid s, l) \right]
\end{eqnarray}
In our implementation, we design a reversible transformation rule that converts a rationale $z$ into a structured, proof-like format $\hat{z}$ and vice versa. For example, given the rationale
\begin{center}
\parbox{0.92\linewidth}{\textit{Response A is mostly helpful... Response B is partially helpful... Thus, Response A is preferred.
}}
\end{center}
we reformulate it into a standardized textual proof: 
\begin{center}
\parbox{0.92\linewidth}{\textit{Here is my justification for why the selected response is the better one. First, Response A is mostly helpful... Response B is partially helpful...
}}
\end{center}
A complete example is provided as shown in the Appendix. Notably, training the preference-proving model requires significantly less annotated data than training the reward model itself, as it only involves teaching the model to explain existing preference judgments rather than learning the preferences from scratch. As a result, a small amount of labeled data is sufficient to train an effective model, as demonstrated empirically in Appendix D.

\paragraph{Preference Proof Selection.}
To enhance the quality and reliability of the generated proofs, we do not rely on a single output from the preference-proving model. Instead, for each input tuple $(s, l)$, we employ a sampling-and-filtering strategy. First, we generate $k$ candidate proofs $\{\hat{z}^{1}, \hat{z}^{2}, \cdots, \hat{z}^{k}\}$ by sampling from the model $\pi_{\psi}$ with a non-zero temperature. We then re-rank the sampled proofs using a probabilistic scoring function. For each candidate $\hat{z}^i$, we compute
\begin{eqnarray}
    \text{Score}(s, l, \hat{z}^i) &=&  - \frac{ \log \pi_{\psi}(\hat{z}^i | s, l)}{ \log \pi_{\psi}(\hat{z}^i)}
\end{eqnarray}
This scoring function produces values in the range $(-\infty, 0]$, with higher scores indicating higher-quality proofs. The basic intuition behind this design is to favour proofs that are highly \textit{specific} to the given context $(s, l)$. It accomplishes this by rewarding proofs that are probable given the context but improbable in isolation, thereby penalizing generic or templated statements that lack contextual relevance. We also provide a theoretical motivation for this approach from a Bayesian perspective in Appendix A.
% In essence, the score approximates the pointwise mutual information (PMI) between the rationale and the conditioning context, encouraging the selection of proofs that are both fluent and discriminative.

\subsection{Self-Training with Unlabeled Data}
The preference-proving model allows us to synthesize rationales for existing labeled data, thereby creating a dataset suitable for training reasoning reward models. However, the performance of this approach is ultimately constrained by the scarcity of the initial labeled preference data. To overcome this bottleneck and further enhance the model's reward reasoning capabilities, we introduce a self-training approach that leverages abundant unlabeled data.

\begin{table*}[!t]
    \centering
    \resizebox{1.0\linewidth}{!}{
    \begin{tabular}{lrcccccccccc}
\toprule[1.1pt]
\multirow{2}{*}{Model} & \multirow{2}{*}{Params.} & \multicolumn{5}{c}{RM-Bench}          & \multicolumn{5}{c}{JudgeBench}  \\ \cmidrule(l){3-7} \cmidrule(l){8-12}
&  & Chat & Math & Code & Safety & Overall & Knowl. & Reason. & Math & Coding & Overall \\  \midrule
\multicolumn{8}{l}{\textbf{\textit{LLM-as-a-Judge}}} \\
GPT-4o$^\sharp$$^\dagger$  &-  &67.2& 67.5& 63.6& 91.7& 72.5 &50.6& 54.1& 75.0& 59.5&59.8  \\
Claude-3.5-Sonnet$^\sharp$$^\dagger$  &-  & 62.5& 62.6& 54.4& 64.4& 61.0 &62.3& 66.3& 66.1& 64.3& 64.8  \\
DeepSeek-R1-0528$^\dagger$ &671B  &76.7 & 74.3& 51.0& 89.2& 72.8 &59.1& 82.7& 80.4& 92.9& 78.8   \\ \midrule
\multicolumn{8}{l}{\textbf{\textit{Open-Source Reward Models}}} \\
Llama-3.1-Nemotron-70B-Reward$^\ddagger$ &70B & 70.7& 64.3& 57.4& 90.3 &70.7 &62.3& 72.5& 76.8& 57.1& 67.2 \\
Skywork-Reward-Gemma-2-27B$^\ddagger$  &27B  & 71.8& 59.2& 56.6& 94.3 &70.5 &59.7& 66.3& 83.9& 50.0& 65.0  \\ 
Skywork-Reward-Llama-3.1-8B$^\ddagger$ &8B &69.5& 60.6& 54.5& 95.7&70.1 &59.1 & 64.3& 76.8& 50.0& 62.5 \\ 
\hdashline
Nemotron-Super$^\ddagger$ & 49B  & 73.7& 91.4& 75.0& 90.6& 82.7 & 71.4& 73.5& 87.5& 76.2& 77.2  \\
Nemotron-Super-Multilingual$^\ddagger$ & 49B  & 77.2& 91.9& 74.7& 92.9& 84.2 & 64.9& 74.5& 87.5& 73.8& 75.2  \\ \midrule
\multicolumn{8}{l}{\textbf{\textit{Reasoning Reward Models}}} \\
RM-R1-Distilled-Qwen-32B &32B  & 74.2& 91.8& 74.1& 95.4& 83.9 & 76.0& 80.6& 88.1& 70.5& 78.8    \\ 
RM-R1-Distilled-Qwen-14B &14B  & 71.8& 90.5& 69.5& 94.1& 81.5 & 68.1 & 72.4 & 87.8 & 84.2 & 78.1   \\ 
RRM-32B     &32B   & 66.6& 81.4& 65.2& 79.4& 73.1  &79.9 & 70.4 & 87.5 & 65.0 & 75.7 \\  \midrule
\multicolumn{9}{l}{\textbf{\textit{Training with Unlabeled Preference Data}}} \\ 
% POLAR-7B  &7B   &  &  &  &  &  &  &  &  &  &   \\ 
% \hdashline
GRAM-Qwen3-14B   &14B &67.4& 55.2& 62.8& 94.3& 69.9  & 63.0& 64.3& 89.3& 69.1& 71.4  \\ 
GRAM-Qwen3-8B   &8B & 63.5& 53.9& 62.9& 92.8& 68.3 & 62.3& 64.3& 80.4& 64.3& 67.8  \\
\midrule

\multicolumn{12}{l}{\textbf{\textit{Training on the Same Labeled Preference Data (LLaMA-3.1-8B-Instruct)}}} \\  
Discriminative RM   &8B  & 70.2& 78.3& 70.1& 85.4& 76.0 & 88.2& 67.1& 85.3& 56.9& 74.4 \\ 
\hdashline
Generative RM     &8B  & 74.8& 81.1& 72.5& 88.6& 79.2 &90.8& 69.4& 87.5& 59.8& 76.9 \\ 
GRAM-R$^2$ (Ours)  &8B  &76.0& 89.8& 80.6& 96.2& 85.7 & 90.9& 83.7& 87.5& 61.9& 81.0 \\ 
\ \ \ +\textit{voting@16} &8B  &\textbf{76.3}& \textbf{90.4}& \textbf{81.2}& \textbf{96.4}& \textbf{86.1}& \textbf{91.2}& \textbf{84.3}& \textbf{88.1}& \textbf{62.8}& \textbf{81.6}  \\ 
\midrule

\multicolumn{12}{l}{\textbf{\textit{Training on the Same Labeled Preference Data (LLaMA-3.2-3B-Instruct)}}} \\
Discriminative RM   &3B  & 70.5& 70.6& 65.5& 95.7& 75.6 &86.0& 70.9& 73.5& 63.2& 73.4  \\ 
\hdashline
Generative RM     &3B  &72.3& 72.1& 68.2& \textbf{95.9}& 77.1 & 90.4& 74.3& 77.4& 64.3& 76.6 \\ 
GRAM-R$^2$ (Ours) &3B  & 74.4& 88.8& 76.6& 95.5& 83.8 &93.0& 78.1& 81.6& 68.5& 80.3    \\
\ \ \ +\textit{voting@16} &3B  & \textbf{74.8}& \textbf{89.4}& \textbf{78.4}& 95.7& \textbf{84.6} &\textbf{93.5}& \textbf{78.6}& \textbf{82.1}& \textbf{69.0}& \textbf{80.8}   \\
\bottomrule[1.1pt]
\end{tabular}}
    \vspace{-0.2cm}
    \caption{
    Accuracies (\%) on RM-Bench and JudgeBench. The best result in each group is in \textbf{bold}. 
    % and the second-best is \underline{underlined}. 
    Results marked with $\sharp$ on RM-Bench are from \citet{chen2025rm}, those with $\dagger$ on JudgeBench are from \citet{liu2025skywork}, and those with $\ddagger$ for both RM-Bench and JudgeBench are from \citet{wang2025helpsteer3}. The other baseline results are either reproduced from their original papers or obtained by evaluating their publicly available models or API access. We use a dotted line to distinguish between the discriminative and generative reward models. 
    }
    \vspace{-0.2cm}
    \label{tab:pair-wise-ranking-res}
\end{table*}

\paragraph{Iterative Self-Training.}
Starting with an initial generative reward model trained on labeled data with synthesized rationales, we iteratively enhance it using batches of unlabeled data $\{D^1_{u}, D^2_{u}, \cdots, D^m_{u}\}$. In the $i$-th iteration, the model is first used to generate preference labels (\textit{i.e.}, preference predictions) for the unlabeled data in batch $D^{i}_{u}$. These pseudo-labeled samples are then fed into the preference-proving model, which synthesizes corresponding rationales. The resulting rationale-based data is merged with the existing synthesized data, and the reward model is retrained on this combined set. This updated model is then used in the next iteration to further improve reward reasoning capabilities. 

\paragraph{Preference Label Denoising.}
A principal challenge in self-training is the propagation of errors from noisy pseudo-labels, which can degrade model performance over successive iterations \cite{xie2020self,das2023understanding}. To mitigate this risk, we implement a multi-pronged denoising strategy that filters both unreliable preference labels and low-quality rationales. First, to enhance label stability, we aggregate predictions from multiple inference runs and apply a majority vote strategy. Second, we enforce a confidence threshold, discarding any pseudo-label whose softmax probability falls below a predefined value. Finally, we validate the rationales themselves through rule-based checks to remove malformed or irrelevant examples. Specifically, we discard examples that contain excessively long rationales, omit rationales to the predicted preference label, or fail to adhere to the structural constraints specified in the prompt.

It is worth noting that a key design choice in our self-training pipeline is the use of a dedicated preference-proving model to generate rationales, rather than relying on those produced internally by the reward model itself. This decision is motivated by the pursuit of high-quality, reliable proofs. While the reward model is trained to perform both reasoning and prediction, the preference-proving model specializes in a single task: generating compelling and coherent proofs. We hypothesize that this specialization provides the preference-proving model with a significant advantage in producing rationales. To validate this hypothesis, we give a comparative experiment in Appendix D.

\section{Experiments}
We evaluate GRAM-R$^2$ on various applications, including response ranking accuracy, adaptability to various reward tasks, and effectiveness in reward-based fine-tuning.

\subsection{Experimental Setups}
\paragraph{Model Backbones.}
For our main experiments, we initialized the preference-proving model with Qwen3-14B \cite{yang2025qwen3}. For the GRAM-R$^2$ model itself, we developed and evaluated two separate versions based on the LLaMA-3.1-8B-Instruct and LLaMA-3.2-3B-Instruct models \cite{dubey2024llama}. The impact of the backbone choice for the preference-proving component is further analyzed in an ablation study in Appendix D.

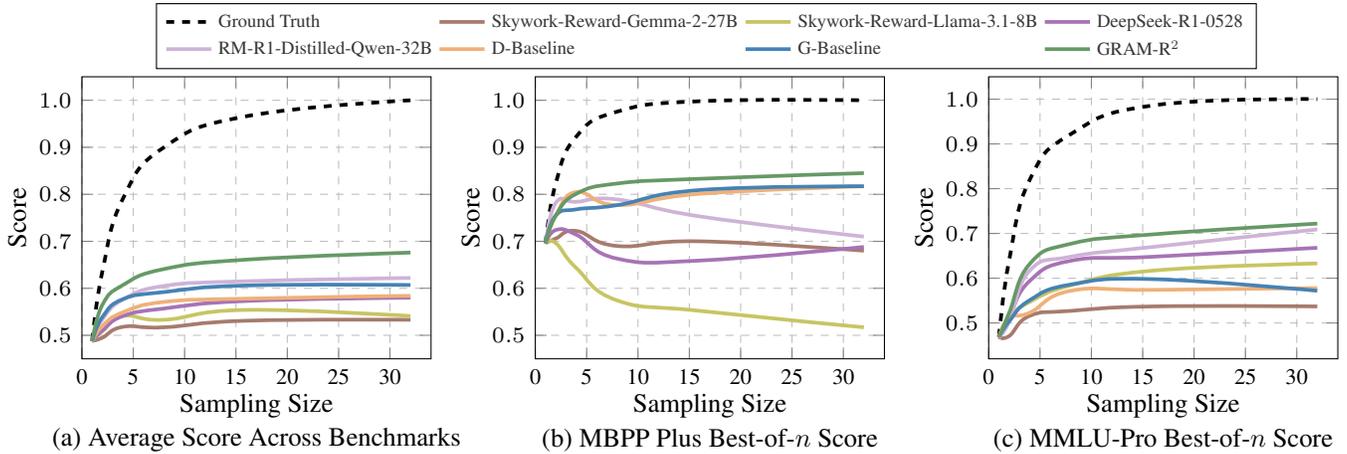
\begin{figure*}[!t]
    \centering
    % \definecolor{green}{RGB}{182,215,168}
% \definecolor{purple}{RGB}{157,193,230}
% \definecolor{blue}{RGB}{135,206,250}
% \definecolor{red}{RGB}{240,128,128}
\definecolor{green}{RGB}{102,153,102}  % green
\definecolor{purple}{RGB}{167,115,181}   % Darker version of the original purple
\definecolor{blue}{RGB}{70,130,180}   % blue
\definecolor{red}{RGB}{239,177,121}    % red
\definecolor{yellow}{RGB}{180,68,62}
\definecolor{brawn}{RGB}{166,123,110}
\definecolor{red2}{RGB}{207,179,215}
\definecolor{oran}{RGB}{199,197,99}

\begin{tikzpicture}
  \hspace{-1.0mm}
  % fig5
  \scriptsize{
    \begin{axis}[
    at={(-6em,-30em)},
    anchor=south west,
    ymajorgrids,
    xmajorgrids,
    grid style=dashed,
    width=.35\textwidth,
    height=.30\textwidth,
    xlabel={\normalsize Sampling Size},
    ylabel={\normalsize Score},
    ylabel style={yshift=-1.5em},
    xlabel style={yshift=0.6em},
    yticklabel style={/pgf/number format/fixed,/pgf/number format/fixed zerofill,/pgf/number format/precision=1,rotate=0,scale=1.0},
    ymin=0.45,
    ymax=1.05, 
    ytick={0.5,0.6,...,1.0},
    xmin=0,
    xmax=34,
    xtick={0,5,...,30},
    xticklabels={0,5,10,15,20,25,30},  % 设置不等间隔的标签
    scaled ticks=false,
    x tick label style={
         anchor=center,
         scale=1.2,
         yshift=-0.8em
     },    
    y tick label style={
         anchor=east,
         scale=1.2,
         xshift=-.1em,
     }, 
    legend style={yshift=30pt,xshift=45em,font={\normalsize},fill opacity=0.8,cells={anchor=west}, legend columns=4},
    title={(a) Average Score Across Benchmarks},
    title style={yshift=-21.5em, font=\normalsize},
    ]
    \addplot[smooth, tension=0.8, black, dashed, mark=none, line width=1.4pt] coordinates {
    (1, 0.488)
    (2, 0.637)
    (4, 0.791)
    (8, 0.900)
    (16, 0.966)
    (32, 1.00)
    };
    % RM 1
    \addlegendentry{\scalebox{.7}{Ground Truth}};
    
    \addplot[smooth, tension=0.8, brawn, mark=none, line width=1.4pt] coordinates {
    (1, 0.488)
    (2, 0.495)
    (4, 0.518)
    (8, 0.517)
    (16, 0.531)
    (32, 0.533)
    };
    % RM 2
    \addlegendentry{\scalebox{.7}{Skywork-Reward-Gemma-2-27B}}

    \addplot[smooth, tension=0.8, oran, mark=none, line width=1.4pt] coordinates {
    (1, 0.488)
    (2, 0.509)
    (4, 0.541)
    (8, 0.533)
    (16, 0.554)
    (32, 0.541)
    };
    % RM 3
    \addlegendentry{\scalebox{.7}{Skywork-Reward-Llama-3.1-8B}}

    \addplot[smooth, tension=0.8, purple, mark=none, line width=1.4pt] coordinates {
    (1, 0.488)
    (2, 0.510)
    (4, 0.541)
    (8, 0.557)
    (16, 0.573)
    (32, 0.580)
    };
    % RM 4
    \addlegendentry{\scalebox{.7}{DeepSeek-R1-0528}}

    \addplot[smooth, tension=0.8, red2, mark=none, line width=1.4pt] coordinates {
    (1, 0.488)
    (2, 0.534)
    (4, 0.576)
    (8, 0.605)
    (16, 0.615)
    (32, 0.622)
    };
    % RM 5
    \addlegendentry{\scalebox{.7}{RM-R1-Distilled-Qwen-32B}}

    \addplot[smooth, tension=0.8, red, mark=none, line width=1.4pt] coordinates {
    (1, 0.488)
    (2, 0.520)
    (4, 0.549)
    (8, 0.571)
    (16, 0.578)
    (32, 0.584)
    };
    % RM 6
    \addlegendentry{\scalebox{.7}{D-Baseline}}

    \addplot[smooth, tension=0.8, blue, mark=none, line width=1.4pt] coordinates {
    (1, 0.488)
    (2, 0.539)
    (4, 0.577)
    (8, 0.592)
    (16, 0.606)
    (32, 0.607)
    };
    % RM 7
    \addlegendentry{\scalebox{.7}{G-Baseline}}

    \addplot[smooth, tension=0.8, green, mark=none, line width=1.4pt] coordinates {
    (1, 0.488)
    (2, 0.565)
    (4, 0.607)
    (8, 0.641)
    (16, 0.661)
    (32, 0.676)
    };
    % RM 8
    \addlegendentry{\scalebox{.7}{GRAM-R$^2$}}

    \end{axis}}

    % fig6
    \scriptsize{
    \begin{axis}[
    at={(18.5em,-30em)},
    anchor=south west,
    ymajorgrids,
    xmajorgrids,
    grid style=dashed,
    width=.35\textwidth,
    height=.30\textwidth,
    xlabel={\normalsize{Sampling Size}},
    ylabel={\normalsize{Score}},
    ylabel style={yshift=-1.5em},
    xlabel style={yshift=0.6em},
    yticklabel style={/pgf/number format/fixed,/pgf/number format/fixed zerofill,/pgf/number format/precision=1,rotate=0,scale=1.0},
    ymin=0.45,
    ymax=1.05, 
    ytick={0.5,0.6,...,1.0},
    xmin=0,
    xmax=34,
    xtick={0,5,...,30},
    xticklabels={0,5,10,15,20,25,30},  % 设置不等间隔的标签
    scaled ticks=false,
    clip=false,
    x tick label style={
         anchor=center,
         scale=1.2,
         yshift=-0.8em
     },    
    y tick label style={
         anchor=east,
         scale=1.2,
         xshift=-.1em,
     }, 
    title={(b) MBPP Plus Best-of-$n$ Score},
    title style={yshift=-21.5em, font=\normalsize},
    ]

    \addplot[smooth, tension=0.8, black, dashed, mark=none, line width=1.4pt] coordinates {
    (1, 0.696)
    (2, 0.824)
    (4, 0.923)
    (8, 0.976)
    (16, 0.998)
    (32, 1.00)
    };
    % RM 1
    \addplot[smooth, tension=0.8, brawn, mark=none, line width=1.4pt] coordinates {
    (1, 0.696)
    (2, 0.706)
    (4, 0.722)
    (8, 0.690)
    (16, 0.700)
    (32, 0.680)
    };
    % RM 2
    \addplot[smooth, tension=0.8, oran, mark=none, line width=1.4pt] coordinates {
    (1, 0.696)
    (2, 0.696)
    (4, 0.643)
    (8, 0.574)
    (16, 0.552)
    (32, 0.517)
    };
    % RM 3
    \addplot[smooth, tension=0.8, purple, mark=none, line width=1.4pt] coordinates {
    (1, 0.696)
    (2, 0.724)
    (4, 0.714)
    (8, 0.663)
    (16, 0.659)
    (32, 0.688)
    };
    % RM 4
    \addplot[smooth, tension=0.8, red2, mark=none, line width=1.4pt] coordinates {
    (1, 0.696)
    (2, 0.782)
    (4, 0.784)
    (8, 0.789)
    (16, 0.753)
    (32, 0.710)
    };
    % RM 5
    \addplot[smooth, tension=0.8, red, mark=none, line width=1.4pt] coordinates {
    (1, 0.696)
    (2, 0.755)
    (4, 0.805)
    (8, 0.777)
    (16, 0.801)
    (32, 0.817)
    };
    % RM 6
    \addplot[smooth, tension=0.8, blue, mark=none, line width=1.4pt] coordinates {
    (1, 0.696)
    (2, 0.755)
    (4, 0.768)
    (8, 0.777)
    (16, 0.809)
    (32, 0.817)
    };
    % RM 7
    \addplot[smooth, tension=0.8, green, mark=none, line width=1.4pt] coordinates {
    (1, 0.696)
    (2, 0.755)
    (4, 0.801)
    (8, 0.823)
    (16, 0.833)
    (32, 0.845)
    };
    % RM 8
    
    \end{axis}  
   }
    % \node [anchor=center] at (.0\textwidth+18.em,-21.6em) {\scalebox{1.0}{(f) RL (LLaMA-3.1-8B-Instruct)}};
  % fig7
  \scriptsize{
    \begin{axis}[
   at={(43em,-30em)},
   anchor=south west,
    ymajorgrids,
    xmajorgrids,
    grid style=dashed,
    width=.35\textwidth,
    height=.30\textwidth,
    xlabel={\normalsize{Sampling Size}},
    ylabel={\normalsize{Score}},
    ylabel style={yshift=-1.5em},
    xlabel style={yshift=0.6em},
    yticklabel style={/pgf/number format/fixed,/pgf/number format/fixed zerofill,/pgf/number format/precision=1,rotate=0,scale=1.0},
    ymin=0.42,
    ymax=1.05, 
    ytick={0.5,0.6,...,1.0},
    xmin=0,
    xmax=34,
    xtick={0,5,...,30},
    xticklabels={0,5,10,15,20,25,30},  % 设置不等间隔的标签
    x tick label style={
         anchor=center,
         scale=1.2,
         yshift=-0.8em
     },    
    y tick label style={
         anchor=east,
         scale=1.2,
         xshift=-.1em,
     }, 
    title={(c) MMLU-Pro Best-of-$n$ Score},
    title style={yshift=-21.5em, font=\normalsize},
    ]
    \addplot[smooth, tension=0.8, black, dashed, mark=none, line width=1.4pt] coordinates {
    (1, 0.469)
    (2, 0.641)
    (4, 0.820)
    (8, 0.922)
    (16, 0.986)
    (32, 1.00)
    };
    % RM 1
    \addplot[smooth, tension=0.8, brawn, mark=none, line width=1.4pt] coordinates {
    (1, 0.469)
    (2, 0.471)
    (4, 0.516)
    (8, 0.527)
    (16, 0.537)
    (32, 0.537)
    };
    % RM 2
    \addplot[smooth, tension=0.8, oran, mark=none, line width=1.4pt] coordinates {
    (1, 0.469)
    (2, 0.512)
    (4, 0.547)
    (8, 0.584)
    (16, 0.617)
    (32, 0.633)
    };
    % RM 3
    \addplot[smooth, tension=0.8, purple, mark=none, line width=1.4pt] coordinates {
    (1, 0.469)
    (2, 0.514)
    (4, 0.594)
    (8, 0.639)
    (16, 0.648)
    (32, 0.668)
    };
    % RM 4
    \addplot[smooth, tension=0.8, red2, mark=none, line width=1.4pt] coordinates {
    (1, 0.469)
    (2, 0.503)
    (4, 0.617)
    (8, 0.648)
    (16, 0.670)
    (32, 0.709)
    };
    % RM 5
    \addplot[smooth, tension=0.8, red, mark=none, line width=1.4pt] coordinates {
    (1, 0.469)
    (2, 0.511)
    (4, 0.523)
    (8, 0.572)
    (16, 0.574)
    (32, 0.578)
    };
    % RM 6
    \addplot[smooth, tension=0.8, blue, mark=none, line width=1.4pt] coordinates {
    (1, 0.469)
    (2, 0.504)
    (4, 0.551)
    (8, 0.586)
    (16, 0.598)
    (32, 0.572)
    };
    % RM 7
    \addplot[smooth, tension=0.8, green, mark=none, line width=1.4pt] coordinates {
    (1, 0.469)
    (2, 0.523)
    (4, 0.632)
    (8, 0.677)
    (16, 0.698)
    (32, 0.722)
    };
    % RM 8

    \end{axis}
   }

\end{tikzpicture}
    \vspace{-0.65cm}
    \caption{
    Best-of-$n$ sampling performance curves for GRAM-R$^2$ and strong baseline models on the PPE benchmark. ``D-Baseline'' and ``G-Baseline'' refer to discriminative and generative reward models, respectively, trained on the same labeled preference data. ``Ground Truth'' represents an oracle reward model that selects responses based on gold-truth answers. All results are reported using the LLaMA-3.1-8B-Instruct backbone. 
    }
    \vspace{-0.2cm}
    \label{fig:list_wise_ranking}
\end{figure*}

\paragraph{Training Datasets.}
Our preference-proving model was trained on the HelpSteer3 dataset \cite{wang2025helpsteer3}, which comprises 40.5K labeled preference examples. Each example was enriched with human-written feedback and a comparative analysis, and we treated this combination as the rationale. For the initial training of GRAM-R$^2$, we curated a 1M-sample rationale-free dataset by amalgamating data from various open sources: MultiPref \cite{miranda2024hybrid}, CodeUltraFeedback \cite{weyssow2024codeultrafeedback}, Unified-Feedback\footnote{https:/huggingface.co/datasets/llm-blender/Unified-Feedback}, Prometheus2-Preference \cite{kim2024prometheus}, PKU-SafeRLHF \cite{ji2023beavertails}, and Skywork-Reward-Preference-80K-v0.2 \cite{liu2024skywork}. The unlabeled data for self-training was sourced from the Stack-Exchange dataset\footnote{https:/huggingface.co/datasets/habedi/stack-exchange-dataset}. To further enhance the model’s reasoning capabilities after pre-training, we performed a fine-tuning step on the human-annotated rationale-based HelpSteer3 dataset. Additional implementation details, including data preprocessing procedures and complete experimental settings, are provided in Appendix B.

\paragraph{Baselines.}
Our primary baselines included strong open-source reward reasoning models, such as RRM \cite{guo2025reward} and RM-R1 \cite{chen2025rm}.  
We also compared GRAM-R$^2$ with several strong baselines: \textit{LLM-as-a-Judge},  where we prompted LLMs like GPT-4o and DeepSeek-V3 to generate preferences; \textit{open-source reward models}, open-source discriminative and generative reward models, including Nemotron-Super-GenRM \cite{wang2025helpsteer3}, and others; and \textit{training on the same labeled preference data}, denoting the standard reward models trained on discriminative and generative frameworks using our labeled preference data, respectively (denoted as Discriminative RM and Generative RM). Furthermore, we compared GRAM-R$^2$ with several approaches designed to utilize the unlabeled data to enhance reward models. These include GRAM, which pre-trains a generative reward model on a response generation task \cite{wang2025gram}. Note that the POLAR model is excluded from this comparison \cite{dou2025pre}, as it requires reference responses not available in these benchmarks. 

\subsection{Pair-wise Response Ranking}
\paragraph{Task Setups.}
Pairwise response ranking is the most commonly used evaluation protocol for reward models. Given an input $x^t$ and two candidate responses, $y^t_a$ and $y^t_b$, the task is to predict the preferred response. Evaluation is conducted on a test set $D^t_{\text{pair}} = {(x^t, y^t_a, y^t_b, l^t)}$, where $l^t$ denotes the ground-truth preference label. Model performance is measured by the accuracy of its predictions against these labels. For this task, we evaluate GRAM-R$^2$ on two widely adopted benchmarks: RM-Bench \cite{liu2024rm}, which assesses the model’s ability to detect subtle stylistic preferences, and JudgeBench \cite{judgebench2024}, which is designed to evaluate generative reward models across diverse tasks.

\paragraph{Results.}
We evaluated the reward reasoning capabilities of GRAM-R$^2$ using the pairwise response ranking task. Table~\ref{tab:pair-wise-ranking-res} reports the performance of GRAM-R$^2$ and various baselines on RM-Bench and JudgeBench. Firstly, a key finding from the results is the consistent and substantial performance improvement brought by incorporating unlabeled data through self-training. Notably, across both backbone settings, GRAM-R$^2$ outperforms both discriminative and generative baselines trained on the same labeled dataset, demonstrating that reward reasoning capabilities can be effectively scaled using large-scale unlabeled data. Furthermore, compared to reasoning reward models that rely on expensive rationale-based annotations or complex reinforcement learning training, GRAM-R$^2$ achieves stronger reward reasoning performance through a simpler and more cost-effective approach, \textit{i.e.}, only using supervised fine-tuning with rationale-free labeled data and unlabeled data. This highlights the practicality and scalability of our approach for training generalist reward models. Additionally, our approach enables the development of compact yet competitive reward models. For instance, our GRAM-R$^2$ model initialized with LLaMA-3.2-3B-Instruct achieves scores of 83.8\% on RM-Bench and 80.3\% on JudgeBench. This performance is remarkably on par with that of the much larger RM-R1-Distilled-Qwen-32B (which scores 83.9\% and 78.8\%, respectively), despite our model being over 10 times smaller.

\begin{figure*}[!t]
    \centering
    \definecolor{green}{RGB}{102,153,102}  % green
\definecolor{purple}{RGB}{167,115,181}   % Darker version of the original purple
\definecolor{blue}{RGB}{70,130,180}   % blue
\definecolor{red}{RGB}{239,177,121}    % red
\definecolor{yellow}{RGB}{180,68,62}
\definecolor{brawn}{RGB}{166,123,110}
\definecolor{red2}{RGB}{207,179,215}
\definecolor{oran}{RGB}{199,197,99}
\definecolor{blue2}{RGB}{145,147,180}

\begin{tikzpicture}  
  % fig1
  \scriptsize{
    \begin{axis}[
    at={(-8em,-18em)},
    anchor=south west,
    ymajorgrids,
    xmajorgrids,
    grid style=dashed,
    width=.28\textwidth,
    height=.24\textwidth,
    xlabel={Amount of Task-Specific Data},
    ylabel={Accuracy (\%) on STEM},
    ylabel style={scale=1.1,yshift=-2.0em},
    xlabel style={scale=1.1,yshift=0.75em},
    yticklabel style={/pgf/number format/fixed,/pgf/number format/fixed zerofill,/pgf/number format/precision=0,rotate=0,scale=1.0,anchor=east,align=right},
    ymin=64,
    ymax=91, 
    ytick={65,70,75,...,95},
    xmin=-0.1,
    xmax=4.1,
    xtick={0,1,2,3,4},
    xticklabels={0K,1K,2K,3K,4K},  
    % 设置不等间隔的标签
    scaled ticks=false,
    x tick label style={
         anchor=center,
         scale=1.2,
         yshift=-0.8em
     },    
    y tick label style={
         anchor=east,
         scale=1.1,
         xshift=-.1em,
     }, 
    legend style={yshift=22pt,xshift=48.5em,font={\small},cells={anchor=west},fill opacity=0.8,legend columns=-1}
    ]

    \addplot[blue2, mark=triangle*,mark size=1.1pt, line width=1.2pt] coordinates {
    (1,68.6)
    (2,71.0)
    (3,75.1)
    (4,76.7)
    };
    % RM 1
    \addlegendentry{\scalebox{.8}{G-Vanilla RM}}

    \addplot[red, mark=square*,mark size=1.1pt,line width=1.2pt] coordinates {
    (0,71.8)
    (1,81.6)
    (2,83.7)
    (3,84.9)
    (4,86.1)
    };
    % RM 2
    \addlegendentry{\scalebox{.8}{RM fine-tuned with G-Baseline}}

    \addplot[brawn, mark=x, mark size=1.6pt, line width=1.2pt] coordinates {
    (0,69.4)
    (1,78.8)
    (2,81.2)
    (3,84.3)
    (4,85.7)
    };
    % RM 3
    \addlegendentry{\scalebox{.8}{RM fine-tuned with GRAM}}

    \addplot[green, mark=*,mark size=1.1pt, line width=1.2pt] coordinates {
    (0,74.3)
    (1,85.7)
    (2,87.8)
    (3,89.0)
    (4,89.8)
    };
    % RM 4
    \addlegendentry{\scalebox{.8}{RM fine-tuned with GRAM-R$^2$}}

    \end{axis}} 
    \node [anchor=center] at (0.11\textwidth,-21.8em) {\scalebox{1.2}{(a) Adaptation (LLaMA-3.1-8B-Instruct)}};

  % fig2
  \scriptsize{
    \begin{axis}[
    at={(10em,-18em)},
    anchor=south west,
    ymajorgrids,
    xmajorgrids,
    grid style=dashed,
    width=.28\textwidth,
    height=.24\textwidth,
    xlabel={Amount of Task-Specific Data},
    ylabel={Accuracy (\%) on Code},
    ylabel style={scale=1.1,yshift=-2.0em},
    xlabel style={scale=1.1,yshift=0.75em},
    yticklabel style={/pgf/number format/fixed,/pgf/number format/fixed zerofill,/pgf/number format/precision=0,rotate=0,scale=1.0,anchor=east,align=right},
    ymin=69,
    ymax=96, 
    ytick={65,70,75,...,95},
    xmin=-0.1,
    xmax=4.1,
    xtick={0,1,2,3,4},
    xticklabels={0K,1K,2K,3K,4K},  
    % 设置不等间隔的标签
    scaled ticks=false,
    x tick label style={
         anchor=center,
         scale=1.2,
         yshift=-0.8em
     },    
    y tick label style={
         anchor=east,
         scale=1.1,
         xshift=-.1em,
     }, 
    legend style={yshift=20pt,xshift=47.5em,font={\small},cells={anchor=west},fill opacity=0.8,legend columns=-1}
    ]

    \addplot[blue2, mark=triangle*,mark size=1.1pt, line width=1.2pt] coordinates {
    (1,75.2)
    (2,80.2)
    (3,80.5)
    (4,84.5)
    };
    % RM 1

    \addplot[red, mark=square*,mark size=1.1pt,line width=1.2pt] coordinates {
    (0,77.6)
    (1,86.3)
    (2,84.2)
    (3,85.2)
    (4,86.8)
    };
    % RM 2
    
    \addplot[brawn, mark=x, mark size=1.6pt, line width=1.2pt] coordinates {
    (0,76.5)
    (1,82.1)
    (2,83.4)
    (3,84.8)
    (4,85.0)
    };
    % RM 3

    \addplot[green, mark=*,mark size=1.1pt, line width=1.2pt] coordinates {
    (0,83.6)
    (1,87.2)
    (2,87.9)
    (3,89.0)
    (4,92.7)
    };
    % RM 4

    \end{axis}} 
    % \node [anchor=center] at (.0\textwidth+18.em,-21.6em) {\scalebox{1.0}{(f) RL (LLaMA-3.1-8B-Instruct)}};

  % fig3
  \scriptsize{
    \begin{axis}[
    at={(28em,-18em)},
    anchor=south west,
    ymajorgrids,
    xmajorgrids,
    grid style=dashed,
    width=.28\textwidth,
    height=.24\textwidth,
    xlabel={Amount of Task-Specific Data},
    ylabel={Accuracy (\%) on STEM},
    ylabel style={scale=1.1,yshift=-2.0em},
    xlabel style={scale=1.1,yshift=0.75em},
    yticklabel style={/pgf/number format/fixed,/pgf/number format/fixed zerofill,/pgf/number format/precision=0,rotate=0,scale=1.0,anchor=east,align=right},
    ymin=64,
    ymax=91, 
    ytick={65,70,75,...,95},
    xmin=-0.1,
    xmax=4.1,
    xtick={0,1,2,3,4},
    xticklabels={0K,2K,4K,6K,8K},  
    % 设置不等间隔的标签
    scaled ticks=false,
    x tick label style={
         anchor=center,
         scale=1.2,
         yshift=-0.8em
     },    
    y tick label style={
         anchor=east,
         scale=1.1,
         xshift=-.1em,
     }, 
    legend style={yshift=20pt,xshift=47.5em,font={\small},cells={anchor=west},fill opacity=0.8,legend columns=-1}
    ]

    \addplot[blue2, mark=triangle*,mark size=1.1pt, line width=1.2pt] coordinates {
    (1,66.9)
    (2,71.2)
    (3,73.1)
    (4,74.7)
    };
    % RM 1

    \addplot[red, mark=square*,mark size=1.1pt,line width=1.2pt] coordinates {
    (0,67.2)
    (1,79.3)
    (2,80.4)
    (3,83.2)
    (4,86.4)
    };
    % RM 2
    
    \addplot[brawn, mark=x, mark size=1.6pt, line width=1.2pt] coordinates {
    (0,69.4)
    (1,78.8)
    (2,81.2)
    (3,84.3)
    (4,85.7)
    };
    % RM 3

    \addplot[green, mark=*,mark size=1.1pt, line width=1.2pt] coordinates {
    (0,71.8)
    (1,82.5)
    (2,83.3)
    (3,85.7)
    (4,88.9)
    };
    % RM 4
    \end{axis}} 
    \node [anchor=center] at (0.6\textwidth,-21.8em) {\scalebox{1.2}{(b) Adaptation (LLaMA-3.2-3B-Instruct)}};

  % fig4
  \scriptsize{
    \begin{axis}[
    at={(46em,-18em)},
    anchor=south west,
    ymajorgrids,
    xmajorgrids,
    grid style=dashed,
    width=.28\textwidth,
    height=.24\textwidth,
    xlabel={Amount of Task-Specific Data},
    ylabel={Accuracy (\%) on Code},
    ylabel style={scale=1.1,yshift=-2.0em},
    xlabel style={scale=1.1,yshift=0.75em},
    yticklabel style={/pgf/number format/fixed,/pgf/number format/fixed zerofill,/pgf/number format/precision=0,rotate=0,scale=1.0,anchor=east,align=right},
    ymin=69,
    ymax=96, 
    ytick={65,70,75,...,95},
    xmin=-0.1,
    xmax=4.1,
    xtick={0,1,2,3,4},
    xticklabels={0K,2K,4K,6K,8K},  
    % 设置不等间隔的标签
    scaled ticks=false,
    x tick label style={
         anchor=center,
         scale=1.2,
         yshift=-0.8em
     },    
    y tick label style={
         anchor=east,
         scale=1.1,
         xshift=-.1em,
     }, 
    legend style={yshift=20pt,xshift=47.5em,font={\small},cells={anchor=west},fill opacity=0.8,legend columns=-1}
    ]

   \addplot[blue2, mark=triangle*,mark size=1.1pt, line width=1.2pt] coordinates {
    (1,73.0)
    (2,78.4)
    (3,81.2)
    (4,83.3)
    };
    % RM 1
    
    \addplot[red, mark=square*,mark size=1.1pt,line width=1.2pt] coordinates {
    (0,74.7)
    (1,80.6)
    (2,81.4)
    (3,83.2)
    (4,86.3)
    };
    % RM 2
    
    \addplot[brawn, mark=x, mark size=1.6pt, line width=1.2pt] coordinates {
    (0,76.5)
    (1,82.1)
    (2,83.4)
    (3,84.8)
    (4,85.0)
    };
    % RM 3

    \addplot[green, mark=*,mark size=1.1pt, line width=1.2pt] coordinates {
    (0,82.4)
    (1,86.7)
    (2,88.6)
    (3,90.2)
    (4,91.3)
    };
    % RM 4
    \end{axis}} 
\end{tikzpicture}
    \vspace{-6mm}
    \caption{
    The performance of reward models fine-tuned with varying amounts of task-specific data (STEM and code generation).
    }
    \vspace{-2mm}
    \label{fig:task_adaptation}
\end{figure*}
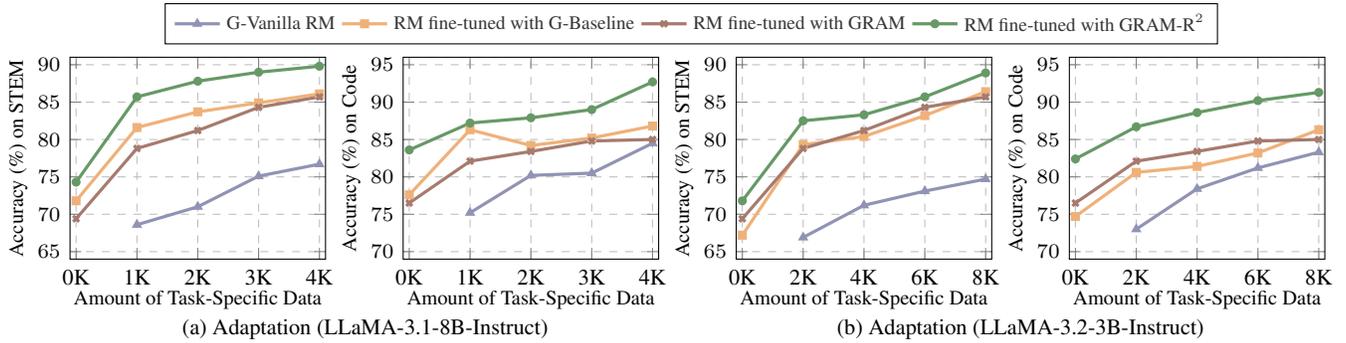

\subsection{List-wise Response Ranking}
\paragraph{Task Setups.}
In practice, multiple candidate responses are typically generated for re-ranking. Given a list-wise test set $D_{\mathrm{list}}^{t} = \{(x^t, y^t_1, y^t_2, \cdots, y^t_n)\}$, where $n$ denotes the number of candidates, the task is to either rank the responses or identify the most preferred one based on human preferences. When the objective is to select the best response, a straightforward strategy involves performing a linear search using the generative reward model. More specifically, we initialize $y^t_b = y^t_1$ as the current best response and iteratively compare it with each remaining candidate. If $y^t_b$ is found to be less preferred during any comparison, it is replaced with the superior response. This process continues until the most preferred response is identified. To improve computational efficiency and support parallelization, we also explore optimized selection algorithms, such as the divide-and-conquer approach. Similarly, this best-response search procedure can be extended to generate a full ranking by repeatedly selecting the best response from the remaining set. Here, to evaluate list-wise ranking performance, we adopt the PPE benchmark \cite{frick2024evaluate}, which includes human preference data from verifiable correctness-based preferences from rigorous benchmarks such as MMLU-Pro and MATH. Specifically, we used the best-of-$n$ (BoN) sampling from PPE to evaluate the ranking quality of our GRAM-R$^2$ model.

\paragraph{Results of Best-of-$n$ Sampling.}
Figure~\ref{fig:list_wise_ranking} presents the BoN sampling performance of GRAM-R$^2$ compared to several strong baselines. A key observation is the prevalence of reward overoptimization \cite{gao2023scaling}, particularly on the MBPP benchmark, where models such as Skywork-Reward-LLaMA-3.1-8B experience significant performance degradation as the number of samples increases. This degradation is primarily due to the limited generalization capabilities of these models to task-specific distributions. In contrast, GRAM-R$^2$ exhibits strong robustness against overoptimization and generalizes effectively across diverse tasks, owing to the incorporation of reward reasoning and self-training on large-scale data. These findings underscore its potential as a reliable reward model for aligning LLMs. Additional evidence is provided in Appendix C, where we show that PPO fine-tuning using GRAM-R$^2$ consistently outperforms PPO fine-tuning using other baselines on the AlpacaEval2 benchmark.

\subsection{Reward Model Adaptation}
We evaluate the adaptability of GRAM-R$^2$ on two distinct tasks: STEM reasoning and code generation. 
% For each task, we fine-tune GRAM-R$^2$ on task-specific, rationale-based labeled data, and then evaluate its performance on the corresponding task-specific test set. 

\paragraph{Task Setups.} 
We randomly sampled STEM and Code task data of varying sizes from the HelpSteer3, using subsets of \{1K, 2K, 3K, 4K\} for STEM and \{2K, 4K, 6K, 8K\} for Code. These subsets are used to fine-tune both GRAM-R$^2$ and its baselines (Generative RM and GRAM-Qwen3-14B). We also trained a generative reward model directly on each dataset as a baseline (\textit{G-Vanilla RM}). All reward models were evaluated on the corresponding held-out validation sets provided by HelpSteer3 for each task.

\paragraph{Results.}
Figure~\ref{fig:task_adaptation} shows the accuracy of reward models fine-tuned on varying amounts of STEM and code data. We observe that GRAM-R$^2$ fine-tunes more effectively into high-quality reward reasoning models compared to training a reward model directly from an LLM backbone. Notably, with 1K STEM samples, GRAM-R$^2$ achieves a task-specific accuracy that exceeds G-Vanilla RM by 17.1 points. GRAM-R$^2$ also consistently outperforms all baselines across various data scales, demonstrating its effectiveness as a foundation reward model that can efficiently adapt to task-specific requirements with minimal supervision.

\begin{figure}[!t]
    \centering
    \vspace{1mm}
    \includegraphics[width=0.98\linewidth]{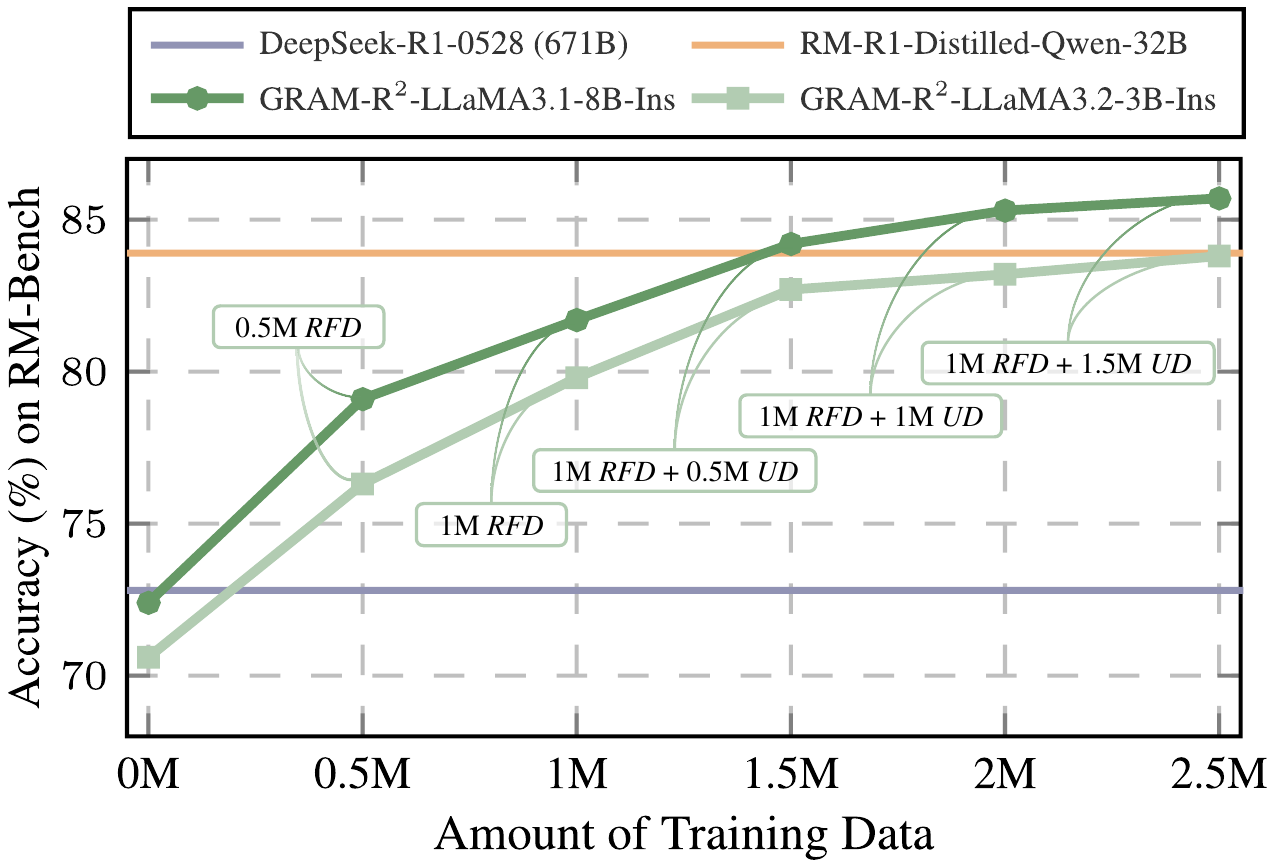}
    \vspace{-2.0mm}
    \caption{
    Performance scaling with different amounts of training data used to pre-train GRAM-R$^2$. ``0M'' denotes the setting where GRAM-R$^2$ is trained solely during the fine-tuning stage, without any pre-training on rationale-free labeled data or unlabeled data. \textit{RFD}: Rationale-Free Labeled Data; \textit{UD}: Unlabeled Data.
    }
    \vspace{-3.5mm}
    \label{fig:gram_scaling}
\end{figure}

\subsection{Analysis}

\paragraph{Scaling Training Data for Improved Performance.}
We explore the impact of training data size on the pre-training performance of GRAM-R$^2$. Specifically, we pre-train GRAM-R$^2$ using datasets of varying sizes: \{0.5M, 1M, 1.5M, 2M, 2.5M\}, each constructed by combining different amounts of rationale-free labeled data and unlabeled data. The model's performance is evaluated on RM-Bench, as shown in Figure~\ref{fig:gram_scaling}.
The results show that increasing the amount of training data generally improves the accuracy of GRAM-R$^2$, with the most notable gains observed when scaling from 0M to 1.5M examples. These findings highlight the importance of both unlabeled data and data scale, suggesting that using both rationale-free labeled data and unlabeled data can substantially enhance the reward reasoning capabilities in reward models.

\section{Conclusions}
We have explored training approaches for reward models with advanced capabilities in reward reasoning. We have developed a generative reward model, called GRAM-R$^2$. The model undergoes initial training on labeled data with synthetic rationales, and then further improves through self-training on large-scale unlabeled data to enhance its reward reasoning capabilities. Extensive experiments demonstrate that GRAM-R$^2$ consistently outperforms various baselines, yielding superior performance in reward reasoning.

\section{Acknowledgments}
This work was supported in part by the National Natural Science Foundation of China (Nos. U24A20334 and 62276056), the Yunnan Fundamental Research Projects (No.202401BC070021), the Yunnan Science and Technology Major Project (No. 202502AD080014), the Fundamental Research Funds for the Central Universities (Nos. N25BSS054 and N25BSS094), and the Program of Introducing Talents of Discipline to Universities, Plan 111 (No.B16009). We would like to thank the anonymous reviewers and SPC for their valuable comments and suggestions that helped improve this paper.

\bibliography{aaai2026}

\clearpage

% \begin{center}
% \Large \textbf{Supplementary Materials for GRAM-R$^2$}
% \end{center}
% \vspace{0.5cm}

\section{Appendix A: Theoretical Motivations of the Preference Proof Selection Approach}
\label{app:motivation_for_preference_pps}
From a Bayesian perspective, selecting the most appropriate proof $\hat{z}$ for a given preference label $l$ over responses $(y_a, y_b)$ under prompt $x$ can be formalized as maximizing the posterior probability $\mathrm{Pr}(\hat{z} \mid s, l)$, where $s = (x, y_a, y_b)$. By Bayes' theorem, the posterior is given by:
\begin{eqnarray}
\mathrm{Pr}(\hat{z} \mid s, l) &=& \frac{\mathrm{Pr}(s, l \mid \hat{z}) \times \mathrm{Pr}(\hat{z})}{\mathrm{Pr}(s, l)}
\end{eqnarray}
Since the marginal likelihood $\mathrm{Pr}(s, l)$ is constant across candidate proofs, the selection objective reduces to maximizing the joint likelihood $\mathrm{Pr}(s, l \mid \hat{z})$ weighted by the prior $\mathrm{Pr}(\hat{z})$. Intuitively, $\mathrm{Pr}(s, l \mid \hat{z})$ quantifies how well a proof explains the given preference label, while $\mathrm{Pr}(\hat{z})$ encodes the generality or plausibility of the proof itself.
In practice, we approximate these distributions using the preference-proving model $\pi_\psi(\cdot)$, treating the model's conditional and unconditional likelihoods as empirical surrogates:
\begin{eqnarray}
\mathrm{Pr}(s, l \mid \hat{z}) \propto \pi_\psi(\hat{z} \mid s, l), \qquad \mathrm{Pr}(\hat{z}) \propto \pi_\psi(\hat{z})
\end{eqnarray}
Substituting these into the posterior and taking logarithms yields the following optimization objective:
\begin{eqnarray}
\hat{z}^* &=& \arg\max_{\hat{z}} \big[ \log \pi_\psi(\hat{z} \mid s, l) - \log \pi_\psi(\hat{z}) \big]
\end{eqnarray}
This corresponds to selecting proofs that are highly likely given the specific context but unlikely under the model's prior, effectively filtering out generic, templated, or overly familiar explanations. In our implementation, we adopt a normalized variant of this expression for scoring, defined as
\begin{eqnarray}
\operatorname{Score}(s, l, \hat{z}) &=& -\frac{\log \pi_\psi(\hat{z} \mid s, l)}{\log \pi_\psi(\hat{z})}
\end{eqnarray}
Since $\log \pi_\psi(\hat{z}) < 0$ in practice, maximizing this score is consistent with the posterior maximization objective above. It can yield high scores for proofs that achieve strong context-conditioned likelihood while being unlikely in isolation, thereby encouraging specificity, informativeness, and contextual relevance. Consequently, under this selection mechanism, the synthesized rationale naturally becomes more dependent on the chosen proof.

Building on this theoretical foundation, our preference proof selection mechanism effectively balances explanatory adequacy and prior plausibility to identify the most credible and contextually grounded proof. Furthermore, the core idea underlying this selection approach has also been validated in recent studies on instruction data selection \cite{li2023quantity,li2024superfiltering}, where they facilitate the selection of more relevant instruction-response pairs, thereby improving the fine-tuning of pre-trained models.

\begin{algorithm}[t]
\caption{GRAM-R2 in Best-of-$n$ Sampling}

\begin{algorithmic}[1]
\Require the input $x$, the candidate responses $\{y_1,\dots,y_n\}$, the trained GRAM-R$^2$ model $\pi_\phi(\cdot)$
\Ensure best response $y_\mathrm{best}$

\State $y_\mathrm{best} \gets y_1$ \Comment{initialize with the first candidate}
\For{$i = 2$ \textbf{to} $n$}
    \State $l \gets \pi_\phi\bigl(x,\, y_\mathrm{best},\, y_i\bigr)$ \Comment{preference label A or B}
    \If{$l = \text{B}$} \Comment{$y_i$ is preferred}
        \State $y_\mathrm{best} \gets y_i$
    \EndIf
\EndFor
\State \Return $y_\mathrm{best}$
\end{algorithmic}
\label{alg:bon}
\end{algorithm}

\section{Appendix B: Details of Experiments}
\label{app:experimental_details}

\subsection{Settings}

\paragraph{Discriminative and Generative Baselines.}
We trained the discriminative and generative reward model baselines for one epoch using a learning rate of 1e-5 and a batch size of 256. For the discriminative baseline, we utilized the complete set of labeled preference data (approximately 1M examples) for training one epoch. As shown in Table~\ref{tab:pair-wise-ranking-res}, this comprehensive training enables our baseline to outperform open-source discriminative reward models such as Skywork-Reward-Llama-3.1-8B, which was trained on only 77K labeled examples. For the generative baseline, we also trained on the complete 1M examples for one epoch, using a learning rate of 3e-6 for LLaMA-3.1-8B-Instruct and 5e-6 for LLaMA-3.2-3B-Instruct. The training template follows the structure illustrated in Figure~\ref{fig:template-g-baseline}. Note that we did not incorporate rationales during training, as the labeled data lacks such annotations.

\paragraph{Preference-Proving Model Training.} 
We trained the preference-proving model for two epochs with a learning rate of 2e-5. During proof generation, we sampled four candidate proofs for each example using top-$p$ sampling, where the $p$ and temperature were set to 0.95 and 0.7, respectively. Our proposed proof selection strategy was then applied to identify the most suitable proof among the candidates, which was subsequently used as the synthesized rationale. During the training, the used template can be found in Figure~\ref{fig:template-preference-proving-model}(a). Additionally, the original HelpSteer3 dataset contains multiple annotations per example, provided by two separate labelers. To unify these dual annotations, we employed GPT-4o to merge the feedback using a template, as shown in Figure~\ref{fig:merge-feedback}.

\begin{figure*}[t]
    \centering
    % \definecolor{green}{RGB}{182,215,168}
% \definecolor{purple}{RGB}{157,193,230}
% \definecolor{blue}{RGB}{135,206,250}
% \definecolor{red}{RGB}{240,128,128}
\definecolor{green}{RGB}{102,153,102}  % green
\definecolor{purple}{RGB}{167,115,181}   % Darker version of the original purple
\definecolor{blue}{RGB}{70,130,180}   % blue
\definecolor{red}{RGB}{239,177,121}    % red
\definecolor{yellow}{RGB}{180,68,62}
\definecolor{brawn}{RGB}{166,123,110}
\definecolor{red2}{RGB}{207,179,215}
\definecolor{oran}{RGB}{199,197,99}

\begin{tikzpicture}
  \hspace{-1.0mm}
    % fig1
    \scriptsize{
    \begin{axis}[
    at={(-6em,-30em)},
    anchor=south west,
    ymajorgrids,
    xmajorgrids,
    grid style=dashed,
    width=.48\textwidth,
    height=.30\textwidth,
    xlabel={\normalsize Training Samples},
    ylabel={\normalsize Oracle Score},
    ylabel style={yshift=-1.5em},
    xlabel style={yshift=0.6em},
    yticklabel style={/pgf/number format/fixed,/pgf/number format/fixed zerofill,/pgf/number format/precision=1,rotate=0,scale=1.0},
    ymin=1.38,
    ymax=2.14, 
    ytick={1.4,1.5,...,2.1},
    xmin=0,
    xmax=21000,
    xtick={0,2000,...,20000},
    xticklabels={0k,2k,4k,6k,8k,10k,12k,14k,16k,18k,20k}, % 设置不等间隔的标签
    scaled ticks=false,
    x tick label style={
         anchor=center,
         scale=1.2,
         yshift=-0.8em
     },    
    y tick label style={
         anchor=east,
         scale=1.2,
         xshift=-.1em,
     }, 
    legend style={yshift=22pt,xshift=32em,font={\normalsize},fill opacity=0.8,cells={anchor=west}, legend columns=-1},
    title={(a) PPO (LLaMA-3.1-8B-Instruct)},
    title style={yshift=-21.5em, font=\normalsize},
    ]
    \addplot[smooth, tension=0.1, brawn, mark=none, line width=1.6pt] coordinates {
    (0.0, 1.42)
    (2560.0, 1.55)
    (5120.0, 1.74)
    (7680.0, 1.72)
    (10240.0, 1.79)
    (12800.0, 1.82)
    (15360.0, 1.85)
    (17920.0, 1.84)
    (20480.0, 1.80)
    };
    % RM 1
    \addlegendentry{\scalebox{.8}{D-Baseline}};
    
    \addplot[smooth, tension=0.1, red2, mark=none, line width=1.6pt] coordinates {
    (0.0, 1.42)
    (2560.0, 1.64)
    (5120.0, 1.80)
    (7680.0, 1.84)
    (10240.0, 1.89)
    (12800.0, 1.92)
    (15360.0, 1.91)
    (17920.0, 1.94)
    (20480.0, 1.95)
    };
    % RM 2
    \addlegendentry{\scalebox{.8}{G-Baseline}}
    
    \addplot[smooth, tension=0.1, oran, mark=none, line width=1.6pt] coordinates {
    (0.0, 1.42)
    (2560.0, 1.65)
    (5120.0, 1.72)
    (7680.0, 1.70)
    (10240.0, 1.74)
    (12800.0, 1.77)
    (15360.0, 1.78)
    (17920.0, 1.75)
    (20480.0, 1.73)
    };
    % RM 3
    \addlegendentry{\scalebox{.8}{GRAM-Qwen3-8B}}
    
    \addplot[smooth, tension=0.1, red, mark=none, line width=1.6pt] coordinates {
    (0.0, 1.42)
    (2560.0, 1.40)
    (5120.0, 1.53)
    (7680.0, 1.62)
    (10240.0, 1.67)
    (12800.0, 1.61)
    (15360.0, 1.60)
    (17920.0, 1.58)
    (20480.0, 1.53)
    };
    % RM 4
    \addlegendentry{\scalebox{.8}{Skywork-Reward-Llama-3.1-8B}}
    
    \addplot[smooth, tension=0.1, green, mark=none, line width=1.6pt] coordinates {
    (0.0, 1.42)
    (2560.0, 1.74)
    (5120.0, 1.85)
    (7680.0, 1.88)
    (10240.0, 2.01)
    (12800.0, 1.96)
    (15360.0, 1.98)
    (17920.0, 2.10)
    (20480.0, 2.08)
    };
    % RM 5
    \addlegendentry{\scalebox{.8}{GRAM-R$^2$}}
    
    \end{axis}} 

    % fig2
    \scriptsize{
    \begin{axis}[
    at={(30em,-30em)},
    anchor=south west,
    ymajorgrids,
    xmajorgrids,
    grid style=dashed,
    width=.48\textwidth,
    height=.30\textwidth,
    xlabel={\normalsize{Training Samples}},
    ylabel={\normalsize{Oracle Score}},
    ylabel style={yshift=-1.5em},
    xlabel style={yshift=0.6em},
    yticklabel style={/pgf/number format/fixed,/pgf/number format/fixed zerofill,/pgf/number format/precision=1,rotate=0,scale=1.0},
    ymin=1.34,
    ymax=2.04, 
    ytick={1.1,1.2,...,2.1},
    xmin=0,
    xmax=21000,
    xtick={0,2000,...,20000},
    xticklabels={0k,2k,4k,6k,8k,10k,12k,14k,16k,18k,20k}, % 设置不等间隔的标签
    scaled ticks=false,
    clip=false,
    x tick label style={
         anchor=center,
         scale=1.2,
         yshift=-0.8em
     },    
    y tick label style={
         anchor=east,
         scale=1.2,
         xshift=-.1em,
     }, 
    title={(b) PPO (LLaMA-3.2-3B-Instruct)},
    title style={yshift=-21.5em, font=\normalsize},
    ]
    \addplot[smooth, tension=0.1, brawn, mark=none, line width=1.6pt] coordinates {
    (0.0, 1.42)
    (2560.0, 1.42)
    (5120.0, 1.46)
    (7680.0, 1.50)
    (10240.0, 1.62)
    (12800.0, 1.64)
    (15360.0, 1.69)
    (17920.0, 1.70)
    (20480.0, 1.68)
    };
    % RM 1
    
    \addplot[smooth, tension=0.1, red2, mark=none, line width=1.6pt] coordinates {
    (0.0, 1.42)
    (2560.0, 1.54)
    (5120.0, 1.58)
    (7680.0, 1.62)
    (10240.0, 1.69)
    (12800.0, 1.74)
    (15360.0, 1.77)
    (17920.0, 1.79)
    (20480.0, 1.75)
    };
    % RM 2

    \addplot[smooth, tension=0.1, oran, mark=none, line width=1.6pt] coordinates {
    (0.0, 1.42)
    (2560.0, 1.65)
    (5120.0, 1.72)
    (7680.0, 1.70)
    (10240.0, 1.74)
    (12800.0, 1.77)
    (15360.0, 1.78)
    (17920.0, 1.75)
    (20480.0, 1.73)
    };
    % RM 3
    
    \addplot[smooth, tension=0.1, red, mark=none, line width=1.6pt] coordinates {
    (0.0, 1.42)
    (2560.0, 1.40)
    (5120.0, 1.53)
    (7680.0, 1.62)
    (10240.0, 1.67)
    (12800.0, 1.61)
    (15360.0, 1.60)
    (17920.0, 1.58)
    (20480.0, 1.53)
    };
    % RM 4

    \addplot[smooth, tension=0.1, green, mark=none, line width=1.6pt] coordinates {
    (0.0, 1.42)
    (2560.0, 1.68)
    (5120.0, 1.76)
    (7680.0, 1.79)
    (10240.0, 1.89)
    (12800.0, 1.95)
    (15360.0, 1.91)
    (17920.0, 1.88)
    (20480.0, 1.98)
    };
    % RM 5
    
    \end{axis}  
    }
    % \node [anchor=center] at (.0\textwidth+18.em,-21.6em) {\scalebox{1.0}{(f) RL (LLaMA-3.1-8B-Instruct)}};
\end{tikzpicture}
    \caption{Results on Reinforcement Learning}
    \label{fig:res-rl}
\end{figure*}
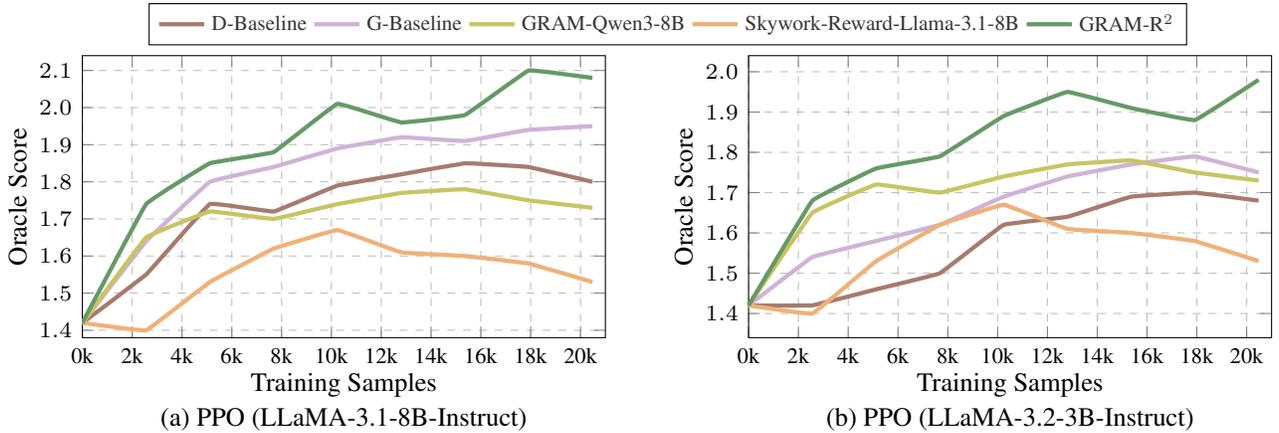

\paragraph{GRAM-R$^2$ Training.} 
In the pre-training stage, we first initialized GRAM-R$^2$ using 1M labeled examples with synthesized rationales. We then performed three iterations of self-training with unlabeled data to enhance the model's reward reasoning capability. During this process, we used a learning rate of 3e-6 for LLaMA-3.1-8B-Instruct and 5e-6 for LLaMA-3.2-3B-Instruct, with the number of training epochs set to one. In each self-training iteration, we began with 0.75M unlabeled examples and applied both format-based and confidence-based filtering to retain a final set of 0.5M high-quality examples. Specifically, we first removed samples that exceeded 4096 tokens or produced label predictions that did not conform to the expected output format. From the remaining examples, we then selected the top 0.5M samples with the highest label prediction confidence. Note that self-training was performed only once using the LLaMA-3.1-8B-Instruct model. The resulting 2.5M pre-training samples were then reused to train GRAM-R$^2$ models based on other backbone models. This design follows recent self-training practices \cite{dubey2024llama,huang2025toolace}, where a stronger backbone model is used to generate high-quality training data, which can then be leveraged to improve the performance and generalization of models with weaker backbones. In the fine-tuning stage, we trained the model for one epoch using a learning rate of 1e-6 for LLaMA-3.1-8B-Instruct and 3e-6 for LLaMA-3.2-3B-Instruct. During the pre-training and fine-tuning stages, the used template is shown in Figure~\ref{fig:template-our-gram-rr}.

\paragraph{Best-of-$n$ Sampling.}
During the best-of-$n$ sampling process, we employed a line search strategy to identify the optimal response among $n$ candidates. This procedure is detailed in Algorithm~\ref{alg:bon}. It is worth noting that to improve efficiency and fully leverage GPU parallelism, we can apply a divide-and-conquer search approach combined with batch generation, enabling the preference labels to be generated in a highly parallelized and scalable manner.

\paragraph{PPO Fine-Tuning.}
We trained the LLM using PPO via the \texttt{trlx} implementation\footnote{\url{https://github.com/CarperAI/trlx}}. For all experiments, the learning rate was set to 1e-5 and 5e-6 for the policy model and the value model, respectively. We settled on a batch size of 64 for each PPO step, which consisted of one epoch of gradient steps and four epochs of mini-batch PPO steps. When using GRAM-R$^2$ to compute reward scores, this optimization objective is then defined as:
\begin{eqnarray}
   \mathcal{L}_{\mathrm{PPO}} &=& - \mathbb{E}_{x\sim D_\mathrm{PPO},\hat{y} \sim \pi_{\theta}} \big[ \gamma \times r_{\phi}(x,\hat{y})) \big] \nonumber\\
    &&- \alpha \times \mathbb{D}_{\mathrm{KL}} \big[ \pi_{\theta}(\hat{y}|x)||\pi_{\theta_\mathrm{ref}}(\hat{y}|x) \big]
\end{eqnarray}
where $\gamma$ denotes a scaling factor, $D_\mathrm{PPO}$ denotes the data for PPO fine-tuning, and $\pi_{\theta_{\mathrm{ref}}}$ denote a reference LLM.
We set $\gamma$ to 10 throughout our experiments. To mitigate the over-optimization issue discussed in \citet{gao2023scaling}, we adopted a checkpointing strategy during training. Specifically, model checkpoints were saved every 200 steps and evaluated on the corresponding validation sets, with the checkpoint achieving the highest reward score selected for final use. Following \citet{wang2024hybrid}, we applied a cold-start strategy for PPO to address the instability caused by inaccurate early value estimates: during the first 30 steps of PPO training, only the value model was updated while the policy model remained fixed. Additionally, inspired by \citet{wang2024esrl}, we standardized the reward scores using a moving reward queue that maintained the most recent 1K scores to compute the running mean and variance.
% All experiments were conducted using eight H100 GPUs.

\subsection{Evaluation}
For evaluation, we mainly used RM-Bench \cite{liu2024rm} and JudgeBench \cite{judgebench2024} to assess pair-wise response ranking performance, and PPE \cite{chen2025towards} to evaluate listwise ranking capabilities. RM-Bench and JudgeBench comprise diverse task subsets, such as chat, code, and math, which allow us to comprehensively evaluate the effectiveness of GRAM-R$^2$ across a broad range of downstream scenarios. Additionally, the PPE benchmark includes widely used evaluation datasets for LLMs, such as MMLU and GPQA. These benchmarks enable us to examine whether GRAM-R$^2$ can effectively enhance the performance of LLMs.

\begin{table}[!t]
    \centering
    \resizebox{0.90\linewidth}{!}{
    \begin{tabular}{lcc}
\toprule[1.1pt]
Method          & WinRate & LC-WinRate \\  \midrule
SFT             & 4.56    &3.08            \\  \midrule
PPO Fine-tuning           &         &            \\
\  \  + D-Baseline        &10.22    &7.36         \\
\  \  + G-Baseline        &11.62    &10.24         \\
\  \  + GRAM-Qwen3-8      &12.17 &10.96            \\
\  \  + Skywork-Reward-8B &9.82  &8.03            \\
\  \  + GRAM-R$^2$        &\bf15.62  &\bf13.80            \\
\bottomrule[1.1pt]
\end{tabular}}
    \caption{
    Win rates of models after PPO fine-tuning with GRAM-R$^2$ and its baselines. ``WinRate'' denotes the raw win rate, while ``LC-WinRate'' denotes the length-controlled win rate. ``Skywork-Reward-8B'' denotes the Skywork-Reward-Llama-3.1-8B model.
    }
    \label{tab:evaluation-rl}
\end{table}

\begin{table*}[!t]
    \centering
    \resizebox{\linewidth}{!}{
    \begin{tabular}{lccccl}
\toprule[1.1pt]
\multirow{2}{*}{Variant} &
\multicolumn{1}{c}{\multirow{2}{*}{\begin{tabular}[c]{@{}c@{}}Reward\\ Reasoning\end{tabular}}} &
\multirow{2}{*}{RFD} &
\multicolumn{2}{c}{UD} &
\multirow{2}{*}{Description} \\ \cmidrule(l){4-5}
 & \multicolumn{1}{c}{} &  & w/ PPM & w/o PPM &  \\  \midrule
GRAM-R$^2$-v1 &  &\checkmark &  &\checkmark   & \parbox{12cm}{A variant without reward reasoning, which skips the preference-proving model and directly self-trains on pseudo-labels without generating rationales.}  \\
GRAM-R$^2$-v2 & \checkmark  &\checkmark  &  &\checkmark         &\parbox{12cm}{A variant that does not use the preference-proving model during training on unlabeled data and instead trains with randomly selected rationales generated by GRAM-R$^2$.}  \\
GRAM-R$^2$-v3 & \checkmark  &  &\checkmark &  & \parbox{12cm}{A variant without training on rationale-free labeled data.} \\
GRAM-R$^2$-v4 & \checkmark  &\checkmark &  &   &\parbox{12cm}{A variant without training on unlabeled data.}  \\
\bottomrule[1.1pt]
\end{tabular}}
    \caption{
    GRAM-R$^2$ variants. RFD: Rationale-Free Labeled Data; UD: Unlabeled Data; PPM: Preference-Proving Model.
    }
    \label{tab:gram-rr-variants}
\end{table*}

\section{Appendix C: Additional Experimental Results}
\label{app:additional_experimental_details}
\subsection{Reinforcement Learning}
In reinforcement learning, the reward score is computed for a single input–response pair $(x, y')$, where $y'$ is sampled from the model. Following \citet{wang2025gram}'s work, we compute this reward using our GRAM-R$^2$ model with a reference response. Specifically, we first obtain the reference response $y_{\mathrm{ref}} = \arg\max \pi_{\theta}(\cdot|x)$ via greedy decoding. We then concatenate the context $c$, input $x'$, sampled response $y'$, and reference response $y_{\mathrm{ref}}$ into a single sequence $s' = [c, x', y', y_{\mathrm{ref}}]$. The final reward for the pair $(x', y')$ is defined as the average probability assigned by the generative reward model $\pi_\phi$, indicating that $y'$ is preferred over the reference response $y_{\mathrm{ref}}$. Specifically, if $y'$ is designated as ``Response A'', the reward score can be computed as:
\begin{eqnarray}
r_{\phi}(x', y') &=& \pi_{\phi}(w=\text{A} \mid s')
\label{eq:apply-generative-rm}
\end{eqnarray}
where the reward score lies in the range $[0, 1]$.

\paragraph{Task Setups.}
To evaluate the performance of GRAM in the reinforcement learning setting, we conducted PPO fine-tuning experiments using the Alpaca dataset \cite{taori2023alpaca}, which contains 52K training examples. We followed the data splits provided by AlpacaFarm \cite{dubois2024alpacafarm} for both supervised fine-tuning (SFT) and PPO training. Notably, we used LLaMA-3.1-8B as the policy model, since the SFT and RLHF training processes for LLaMA-3.1-8B-Instruct have not been publicly released. This lack of transparency introduces a data distribution shift that is incompatible with our experimental setup. Following prior work \cite{wang2025gram,yang2024regularizing}, we included an oracle reward score in our evaluation, computed using a discriminative reward model trained on preference data from AlpacaFarm. This oracle model provides an accurate measure of response quality and serves as a tool to assess generalization, as AlpacaFarm's preference data is co-distributed with the AlpacaEval2 test data.

\paragraph{Results of PPO Fine-Tuning.}
We apply GRAM-R$^2$ and its baselines as reward models in PPO fine-tuning. As shown in Figure~\ref{fig:res-rl}, the observed behavior during reinforcement learning is similar to that seen with BoN sampling. For baseline methods, the oracle scores begin to decline early in training, while their corresponding proxy scores continue to rise, indicating a clear overoptimization issue. In contrast, GRAM-R$^2$ exhibits stronger generalization, as reflected in the consistent improvement of the oracle score. These results demonstrate that GRAM-R$^2$ effectively mitigates reward overoptimization during PPO fine-tuning. 
Here, we attribute the superior generalization in GRAM-R$^2$ to two key factors. First, the explicit incorporation of reward reasoning enables the model to provide more reliable reward signals, reducing the risk of overoptimization. Recent studies corroborate this finding \cite{liang2025generative,guo2025reward}. Second, our self-training strategy leverages vast unlabeled data during the pre-training stage, which significantly enhances the model's robustness and generalization ability.

\paragraph{Performance Comparison of LLMs Trained via Different Reward Models.} 
To test its effectiveness in PPO fine-tuning, we further evaluate the performance of an LLM fine-tuned using GRAM-R$^2$ as the reward signal. For comparison, we train separate policies using several strong baseline reward models, including D-Baseline, G-Baseline, GRAM-Qwen3-8, and Skywork-Reward-Llama-3.1-8B. The quality of these fine-tuned LLMs is then benchmarked using the \texttt{alpaca\_eval} system\footnote{https://github.com/tatsu-lab/alpaca\_eval}, where GPT-4 acts as an automated judge to compute the win rate of each model's responses against a standard baseline. As shown in Table~\ref{tab:evaluation-rl}, the LLM trained with GRAM-R$^2$ achieves the highest win rate, demonstrating that it provides a more effective reward signal for guiding PPO fine-tuning.

\begin{figure*}[!t]
    \centering
    \definecolor{green}{RGB}{102,153,102}
\definecolor{bargram_green}{RGB}{182, 215, 168}
\definecolor{purple}{RGB}{167,115,181}   % Darker version of the original purple
\definecolor{blue}{RGB}{70,130,180}   % blue
\definecolor{red}{RGB}{239,177,121}    % red
\definecolor{yellow}{RGB}{180,68,62}
\definecolor{brawn}{RGB}{166,123,110}
\definecolor{red2}{RGB}{207,179,215}
\definecolor{oran}{RGB}{199,197,99}
\definecolor{blue2}{RGB}{145,147,180}
\definecolor{white}{RGB}{255,255,255}
\begin{tikzpicture}
  \scriptsize{
    \begin{axis}[
    enlargelimits={abs=0.5},
    at={(-20em,0)},
    anchor=south west,
    ymajorgrids,
    % xmajorgrids,
    grid style=dashed,
    width=.48\textwidth,
    height=.35\textwidth,
    ybar,
    bar width=13pt,
    nodes near coords,              
    nodes near coords align={vertical}, 
    xtick align=inside,
    every node near coord/.append style={
        /pgf/number format/.cd, 
        fixed, zerofill, 
        precision=1   
    },
    ylabel={\scalebox{1.2}{Accuracy (\%)}},
    ylabel style={yshift=-2em},
    xlabel style={yshift=0.55em},
    yticklabel style={/pgf/number format/precision=1,/pgf/number format/fixed zerofill},
    ymin=63,
    ymax=88, 
    ytick={50,55,...,90},
    xtick={0.5,1.5,2.5,3.5,4.5,5.5},
    xticklabels={GRAM-R$^2$-v1,GRAM-R$^2$-v2,GRAM-R$^2$-v3,GRAM-R$^2$-v4,G-Baseline,GRAM-R$^2$},
    xticklabel style={rotate=23},
    legend style={
        at={(1.7,1.2)},   
        cells={anchor=east},
        legend columns=-1,
        draw=black,
        /tikz/every even column/.append style={column sep=0.5cm},
        % cells={align=left}
    }
    ]
      
    \addplot[fill=bargram_green!60,draw=black] coordinates {
    (0.5,80.4) 
    (1.5,84.1) 
    (2.5,78.8) 
    (3.5,82.6) 
    (4.5,79.2)
    (5.5,85.7)
    };

    \end{axis}
   }
   
    \node [anchor=center] at (-.08\textwidth,-10ex) {\scalebox{1.2}{(a) Performance on Pair-wise Response Ranking}};
  \scriptsize{
    \begin{axis}[
    at={(15em,0)},
    anchor=south west,
    ymajorgrids,
    % xmajorgrids,
    grid style=dashed,
    width=.48\textwidth,
    height=.35\textwidth,
    % ybar,
    % bar width=9.5pt,
    % bar width=1.4em,
    % enlarge x limits=2,
    xtick align=inside,
    xlabel={\scalebox{1.2}{Amount of Preference Data}},
    ylabel={\scalebox{1.2}{Accuracy (\%)}},
    ylabel style={yshift=-1.8em},
    xlabel style={yshift=0.55em},
    yticklabel style={/pgf/number format/precision=1,/pgf/number format/fixed zerofill},
    ymin=53,
    ymax=92,
    xmin=-0.1,
    xmax=4.1,
    ytick={55,60,...,90},
    xtick={0,1,2,3,4},
    xticklabels={0k,2k,4k,6k,8k},
    legend style={
        at={(0.96, 0.56)},   
        cells={anchor=west},
        legend columns=1
        draw=black!10,
        opacity=0.7,
        legend style={cells={align=left}},
        /tikz/every even column/.append style={column sep=0.5cm}
    }
    ]
    
    \addplot[red, mark=square*,mark size=1.5pt,line width=1.0pt] coordinates {
    (0,66.2)
    (1,77.3)
    (2,79.7)
    (3,82.2)
    (4,83.9)
    };
    \addlegendentry{\scalebox{1.}{RM fine-tuned with GRAM-R$^2$-v1}}

    \addplot[blue, mark=square,mark size=2pt,line width=1.0pt] coordinates {
    (0,70.4)
    (1,84.3)
    (2,85.4)
    (3,86.8)
    (4,88.8)
    };
    \addlegendentry{\scalebox{1.}{RM fine-tuned with GRAM-R$^2$-v2}}

    \addplot[red2, mark=o,mark size=2pt,line width=1.0pt] coordinates {
    (0,67.4)
    (1,78.0)
    (2,78.9)
    (3,80.1)
    (4,81.4)
    };
    \addlegendentry{\scalebox{1.}{RM fine-tuned with GRAM-R$^2$-v3}}

    \addplot[yellow, mark=+,mark size=2.5pt,line width=1.0pt] coordinates {
    (0,72.3)
    (1,83.2)
    (2,84.5)
    (3,88.3)
    (4,89.2)
    };
    \addlegendentry{\scalebox{1.}{RM fine-tuned with GRAM-R$^2$-v4}}

    % \addlegendimage{empty legend};
    % \addlegendentry{}; 

    % \addplot[brawn, mark=star,mark size=2.5pt,line width=1.0pt] coordinates {
    % (0,51.3)
    % (1,61.3)
    % (3,68.9)
    % (5,69.2)
    % (7,69.5)
    % (10,69.9)
    % };
    % \addlegendentry{\scalebox{1.}{RM fine-tuned with D-Baseline}}

    \addplot[purple, mark=x,mark size=2.5pt,line width=1.0pt] coordinates {
    (0,71.8)
    (1,81.6)
    (2,83.7)
    (3,84.9)
    (4,86.1)
    };
    \addlegendentry{\scalebox{1.}{RM fine-tuned with G-Baseline}}

     \addplot[green,  mark=*,mark size=1.2pt, line width=1.0pt] coordinates {
    (0,74.3)
    (1,85.7)
    (2,87.8)
    (3,89.0)
    (4,89.8)
    };
    \addlegendentry{\scalebox{1.}{RM fine-tuned with GRAM-R$^2$}}

    \end{axis}  
   }
    \node [anchor=center] at (-.08\textwidth+35.0em,-10ex) {\scalebox{1.2}{(b) Performance on Reward Model Adaptation}};

\end{tikzpicture}
    \caption{
   We employ LLaMA-3.1-8B-Instruct as the backbone model and evaluate different GRAM-R$^2$ variants on pair-wise response ranking using RM-Bench and on reward model adaptation using STEM.
    }
    \label{fig:gram-variants}
\end{figure*}

\subsection{Comparing GRAM-R$^2$ with Generative Baselines}
As shown in Table 1, the standard generative baseline (G-Baseline) achieves impressive accuracy when trained on our 1M-sample labeled data. For instance, the LLaMA-3.1-8B-Instruct version reaches 79.2\% on the RM-Bench, outperforming strong baselines like GPT-4o. However, this strong performance proves to be brittle and does not generalize to other evaluation settings. Specifically, its performance degrades significantly in dynamic, out-of-distribution scenarios such as BoN sampling (Figure~\ref{fig:list_wise_ranking}) and task adaptation (Figure~\ref{fig:task_adaptation}).
We attribute this inconsistency to severe overfitting on the labeled training data. While the model excels on several benchmarks, it lacks the broader generalization required for more complex tasks. In contrast, GRAM-R$^2$ is designed to overcome this limitation. By integrating explicit reward reasoning and training on vast unlabeled data, our model develops superior generalization capabilities, allowing it to maintain strong and consistent performance across different downstream tasks.

\section{Appendix D: More Analysis}
\label{app:analysis}

\subsection{Ablation Study on Self-Training}
To isolate the contribution of each component within our self-training approach, we conduct a detailed ablation study with different GRAM-R$^2$ variants as shown in Table~\ref{tab:gram-rr-variants}. We evaluate these GRAM-R$^2$ variants through experiments on pair-wise response ranking and reward model adaptation. 

The results are summarized in Figure~\ref{fig:gram-variants}. First, comparing GRAM-R$^2$ with GRAM-R$^2$-v1 highlights the critical role of reward reasoning: GRAM-R$^2$ achieves 85.7\% accuracy on RM-Bench, significantly outperforming GRAM-R$^2$-v1 at 80.4\%. This confirms that incorporating reward reasoning is essential for training effective reward models. Second, GRAM-R$^2$-v2 achieves 84.1\% accuracy, demonstrating the effectiveness of using the preference-proving model during self-training. This result supports our central insight: generating rationales via structured proof guidance helps produce higher-quality pseudo-labels and improves generalization. Third, when comparing GRAM-R$^2$-v3 and GRAM-R$^2$-v4, we can observe that pretraining with rationale-free labeled data still provides a significant performance gain, underscoring the importance of high-quality preference annotations even in the absence of explicit reasoning components. Finally, GRAM-R$^2$-v3 delivers strong performance despite relying solely on 1.5M unlabeled examples and no additional labeled data during self-training. Its competitive accuracy of 82.6\% illustrates the potential of unlabeled data in enhancing reward reasoning when combined with a well-designed self-training pipeline.

\subsection{Performance of Preference-Proving Model with Different Backbone Models}
To evaluate the performance of preference-proving models across different backbone architectures, we begin by sampling 100K rationale-free labeled examples. We then train separate preference-proving models using Qwen3-8B, LLaMA-3.1-8B-Instruct, Qwen3-14B, and Qwen3-32B as backbones. Each model is used to generate preference proofs and synthesize corresponding rationales for the sampled data. Additionally, we compare these models with a prompting-based approach, where strong LLMs such as DeepSeek-R1 and GPT-4o are directly prompted to generate proofs and synthesize rationales. Finally, the generated rationales are used to fine-tune an LLaMA-3.1-8B-Instruct model on the resulting synthesized rationales. The results are listed in Table~\ref{tab:ppm_with_diff_models}.
We observe that training the preference-proving model on labeled data consistently leads to better downstream performance. We also find that among the models evaluated, larger backbone models generally yield stronger results. For example, PPMs based on Qwen3-14B and Qwen3-32B outperform those using smaller backbones such as Qwen3-8B or LLaMA-3.1-8B-Instruct. However, the performance gain from Qwen3-14B to Qwen3-32B is marginal (74.3\% vs. 74.8\% on RM-Bench), suggesting diminishing returns with increased model size. Given the computational demands of large-scale self-training, we choose Qwen3-14B as the backbone for our final preference-proving model to reduce overall compute cost while maintaining strong performance.

\begin{table}[!t]
    \centering
    \resizebox{0.92\linewidth}{!}{
    \begin{tabular}{lcc}
\toprule[1.1pt]
Model & RM-Bench & JudgeBench \\ \midrule
GRAM-R$^2$-100k               &          &          \\
\ \ \ w/ PPM-GPT-4o           &64.7          &63.6          \\
\ \ \ w/ PPM-DeeepSeek-R1     &66.2          &65.4          \\
\ \ \ w/ PPM-Qwen3-8B         &69.3          &68.7          \\
\ \ \ w/ PPM-LLaMA-3.1-8B     &72.6          &71.2          \\
\ \ \ w/ PPM-Qwen3-14B        &74.3          &73.5          \\
\ \ \ w/ PPM-Qwen3-32B       &\bf74.8          &\bf74.4          \\
\bottomrule[1.1pt]
\end{tabular}}
    \caption{
   Performance of preference-proving models trained with different backbone architectures. ``-100K'' indicates that GRAM-R$^2$ was trained using only 100K rationale-free labeled examples. PPM: Preference-Proving Model.
    }
    \label{tab:ppm_with_diff_models}
\end{table}

\subsection{Performance of Preference-Proving Model with Different Sampling Sizes}
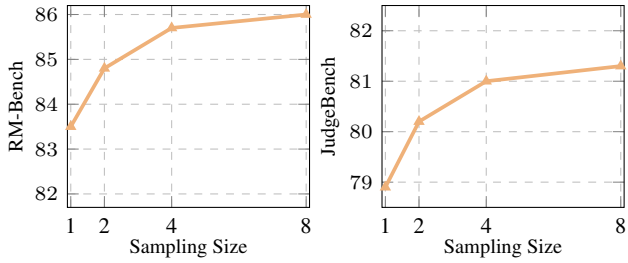
\begin{figure}[!t]
    \centering
    \definecolor{green}{RGB}{102,153,102}  % green
\definecolor{purple}{RGB}{167,115,181}   % Darker version of the original purple
\definecolor{blue}{RGB}{70,130,180}   % blue
\definecolor{red}{RGB}{239,177,121}    % red
\definecolor{yellow}{RGB}{180,68,62}
\definecolor{brawn}{RGB}{166,123,110}
\definecolor{red2}{RGB}{207,179,215}
\definecolor{oran}{RGB}{199,197,99}
\definecolor{blue2}{RGB}{145,147,180}

\begin{tikzpicture}  
  % fig1
  \scriptsize{
    \begin{axis}[
    at={(-9em,-18em)},
    anchor=south west,
    ymajorgrids,
    xmajorgrids,
    grid style=dashed,
    width=.27\textwidth,
    height=.24\textwidth,
    xlabel={Sampling Size},
    ylabel={RM-Bench},
    ylabel style={scale=1.1,yshift=-2.0em},
    xlabel style={scale=1.1,yshift=0.75em},
    yticklabel style={/pgf/number format/fixed,/pgf/number format/fixed zerofill,/pgf/number format/precision=0,rotate=0,scale=1.0,anchor=east,align=right},
    ymin=81.7,
    ymax=86.2, 
    ytick={82,83,84,85,86},
    xmin=0.9,
    xmax=8.1,
    xtick={1,2,4,8},
    xticklabels={1,2,4,8},  
    % 设置不等间隔的标签
    scaled ticks=false,
    x tick label style={
         anchor=center,
         scale=1.2,
         yshift=-0.8em
     },    
    y tick label style={
         anchor=east,
         scale=1.1,
         xshift=-.1em,
     }, 
    legend style={yshift=30pt,xshift=18.2em,font={\small},cells={anchor=west},fill opacity=0.8,legend columns=2}
    ]

    \addplot[red, mark=triangle*,mark size=1.1pt, line width=1.4pt] coordinates {
    (1,83.5)
    (2,84.8)
    (4,85.7)
    (8,86.0)
    };
    \end{axis}} 

  % fig2
  \scriptsize{
    \begin{axis}[
    at={(8em,-18em)},
    anchor=south west,
    ymajorgrids,
    xmajorgrids,
    grid style=dashed,
    width=.27\textwidth,
    height=.24\textwidth,
    xlabel={Sampling Size},
    ylabel={JudgeBench},
    ylabel style={scale=1.1,yshift=-2.0em},
    xlabel style={scale=1.1,yshift=0.75em},
    yticklabel style={/pgf/number format/fixed,/pgf/number format/fixed zerofill,/pgf/number format/precision=0,rotate=0,scale=1.0,anchor=east,align=right},
    ymin=78.5,
    ymax=82.5, 
    ytick={77,78,...,82},
    xmin=0.9,
    xmax=8.1,
    xtick={1,2,4,8},
    xticklabels={1,2,4,8},  
    scaled ticks=false,
    x tick label style={
         anchor=center,
         scale=1.2,
         yshift=-0.8em
     },    
    y tick label style={
         anchor=east,
         scale=1.1,
         xshift=-.1em,
     }, 
    legend style={yshift=20pt,xshift=47.5em,font={\small},cells={anchor=west},fill opacity=0.8,legend columns=-1}
    ]

    \addplot[red, mark=triangle*,mark size=1.1pt, line width=1.4pt] coordinates {
    (1,78.9)
    (2,80.2)
    (4,81.0)
    (8,81.3)
    };
    \end{axis}} 

  % fig3
\end{tikzpicture}
    \vspace{-6mm}
    \caption{Effect of sampling size on the performance of the preference-based proof selection mechanism. }
    \label{fig:ppm_with_diff_sampling_size}
\end{figure}

We investigate the impact of sampling size on the performance of our proof selection mechanism by evaluating sampling sizes of \{1, 2, 4, 8\}. A sampling size of 1 serves as the baseline, corresponding to a scenario without proof selection, where the rationale is synthesized directly from a single, unfiltered generation. As shown in Figure~\ref{fig:ppm_with_diff_sampling_size}, results from the LLaMA-3.1-8B-Instruct model on both RM-Bench (left) and JudgeBench (right) reveal a clear trend: performance is lowest when using a sampling size of 1, confirming the effectiveness of our preference-based proof selection strategy. Moreover, we observe that performance improvements begin to plateau once the sampling size reaches 4, with only marginal gains observed at size 8. This suggests that a sampling size of 4 offers a good balance between performance and computational cost, effectively covering the proof space with diminishing returns beyond that point. 

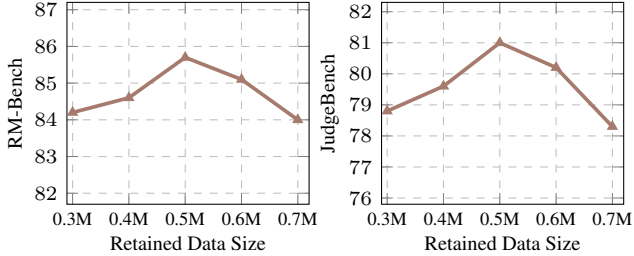
\begin{figure}[!t]
    \centering
    \definecolor{green}{RGB}{102,153,102}  % green
\definecolor{purple}{RGB}{167,115,181}   % Darker version of the original purple
\definecolor{blue}{RGB}{70,130,180}   % blue
\definecolor{red}{RGB}{239,177,121}    % red
\definecolor{yellow}{RGB}{180,68,62}
\definecolor{brawn}{RGB}{166,123,110}
\definecolor{red2}{RGB}{207,179,215}
\definecolor{oran}{RGB}{199,197,99}
\definecolor{blue2}{RGB}{145,147,180}

\begin{tikzpicture}  
  % fig1
  \scriptsize{
    \begin{axis}[
    at={(-9em,-18em)},
    anchor=south west,
    ymajorgrids,
    xmajorgrids,
    grid style=dashed,
    width=.27\textwidth,
    height=.24\textwidth,
    xlabel={Retained Data Size},
    ylabel={RM-Bench},
    ylabel style={scale=1.1,yshift=-2.0em},
    xlabel style={scale=1.1,yshift=0.75em},
    yticklabel style={/pgf/number format/fixed,/pgf/number format/fixed zerofill,/pgf/number format/precision=0,rotate=0,scale=1.0,anchor=east,align=right},
    ymin=81.7,
    ymax=87.2, 
    ytick={82,83,84,85,86,87},
    xmin=0.9,
    xmax=5.2,
    xtick={1,2,3,4,5},
    xticklabels={0.3M,0.4M,0.5M,0.6M,0.7M},  
    % 设置不等间隔的标签
    scaled ticks=false,
    x tick label style={
         anchor=center,
         scale=1.0,
         yshift=-0.8em
     },    
    y tick label style={
         anchor=east,
         scale=1.1,
         xshift=-.1em,
     }, 
    legend style={yshift=30pt,xshift=18.2em,font={\small},cells={anchor=west},fill opacity=0.8,legend columns=2}
    ]

    \addplot[brawn, mark=triangle*,mark size=1.1pt, line width=1.4pt] coordinates {
    (1,84.2)
    (2,84.6)
    (3,85.7)
    (4,85.1)
    (5,84.0)
    };
    \end{axis}} 

  % fig2
  \scriptsize{
    \begin{axis}[
    at={(8em,-18em)},
    anchor=south west,
    ymajorgrids,
    xmajorgrids,
    grid style=dashed,
    width=.27\textwidth,
    height=.24\textwidth,
    xlabel={Retained Data Size},
    ylabel={JudgeBench},
    ylabel style={scale=1.1,yshift=-2.0em},
    xlabel style={scale=1.1,yshift=0.75em},
    yticklabel style={/pgf/number format/fixed,/pgf/number format/fixed zerofill,/pgf/number format/precision=0,rotate=0,scale=1.0,anchor=east,align=right},
    ymin=75.8,
    ymax=82.3, 
    ytick={76,77,...,82},
    xmin=0.9,
    xmax=5.2,
    xtick={1,2,3,4,5},
    xticklabels={0.3M,0.4M,0.5M,0.6M,0.7M},  
    scaled ticks=false,
    x tick label style={
         anchor=center,
        scale=1.0,
         yshift=-0.8em
     },    
    y tick label style={
         anchor=east,
         scale=1.1,
         xshift=-.1em,
     }, 
    legend style={yshift=20pt,xshift=47.5em,font={\small},cells={anchor=west},fill opacity=0.8,legend columns=-1}
    ]

    \addplot[brawn, mark=triangle*,mark size=1.1pt, line width=1.4pt] coordinates {
    (1,78.8)
    (2,79.6)
    (3,81.0)
    (4,80.2)
    (5,78.3)
    };
    \end{axis}} 

  % fig3
\end{tikzpicture}
    \vspace{-6mm}
    \caption{The impact of retained data size on the performance of self-training our GRAM-R$^2$.}
    \vspace{-2mm}
    \label{fig:self_training_diff_data_filter}
\end{figure}

\subsection{Self-Training Performance under Different Data Filtering Sizes}
In our iterative self-training process, to prevent the propagation of erroneous labels, we filter the data based on confidence. Here, we test the impact of the retained data size per round on the GRAM-R$^2$ model with the LLaMA-3.1-8B-Instruct model. As shown in Figure~\ref{fig:self_training_diff_data_filter}, we find a distinct performance peak at a data size of 0.5M on both RM-Bench and JudgeBench. This suggests an optimal trade-off: retaining too little data (\textit{e.g.}, 0.3M) results in an information bottleneck, while retaining too much (\textit{e.g.}, 0.7M) introduces excessive noise from low-confidence pseudo-labels. Consequently, we choose 0.5M as the optimal data size for our filtering strategy.

\subsection{Test-Time Scaling of GRAM-R$^2$}
As a reward reasoning model, our GRAM-R$^2$ possesses the unique capability for test-time scaling. Specifically, we implement this through the straightforward yet effective method of BoN sampling. We evaluate the resulting accuracy improvements on two distinct models: LLaMA-3.1-8B-Instruct and LLaMA-3.2-3B-Instruct. The experimental results, presented in Figure 1, reveal a key advantage of our approach. In contrast to traditional discriminative reward models, GRAM-R² can leverage the inherent scaling properties of generative models at inference time to significantly boost its reward accuracy. This finding not only validates the promise of the reward reasoning paradigm but also suggests a promising future direction where advanced techniques from the broader LLM landscape can be continually adapted to enhance the capabilities of reward models.

\begin{figure}[!t]
    \centering
    \definecolor{green}{RGB}{102,153,102}  % green
\definecolor{purple}{RGB}{167,115,181}   % Darker version of the original purple
\definecolor{blue}{RGB}{70,130,180}   % blue
\definecolor{red}{RGB}{239,177,121}    % red
\definecolor{yellow}{RGB}{180,68,62}
\definecolor{brawn}{RGB}{166,123,110}
\definecolor{red2}{RGB}{207,179,215}
\definecolor{oran}{RGB}{199,197,99}
\definecolor{blue2}{RGB}{145,147,180}

\begin{tikzpicture}
  \hspace{-1mm}
  % fig1
  \scriptsize{
    \begin{axis}[
    at={(-9em,-18em)},
    anchor=south west,
    ymajorgrids,
    xmajorgrids,
    grid style=dashed,
    width=.27\textwidth,
    height=.24\textwidth,
    xlabel={Sampling Size},
    ylabel={RM-Bench},
    ylabel style={scale=1.1,yshift=-2.0em},
    xlabel style={scale=1.1,yshift=0.75em},
    yticklabel style={/pgf/number format/fixed,/pgf/number format/fixed zerofill,/pgf/number format/precision=0,rotate=0,scale=1.0,anchor=east,align=right},
    ymin=82.7,
    ymax=88.5, 
    ytick={80,81,...,90},
    xmin=-0.5,
    xmax=16.5,
    xtick={0,2,4,7,11,16},
    xticklabels={1,8,16,32,64,128},  
    % 设置不等间隔的标签
    scaled ticks=false,
    x tick label style={
         anchor=center,
         scale=1.2,
         yshift=-0.8em
     },    
    y tick label style={
         anchor=east,
         scale=1.1,
         xshift=-.1em,
     }, 
    legend style={yshift=20pt,xshift=17em,font={\small},cells={anchor=west},fill opacity=0.8,legend columns=2}
    ]

    \addplot[green, mark=triangle*,mark size=1.1pt, line width=1.4pt] coordinates {
    (0,85.7)
    (2,86.0)
    (4,86.1)
    (7,86.8)
    (11,87.4)
    (16,87.9)
    };
    % RM 1
    \addlegendentry{\scalebox{.66}{GRAM-R$^2$ (LLaMA-3.1-8B-Ins)}}

    \addplot[green!50, mark=square*,mark size=1.1pt,line width=1.4pt] coordinates {
    (0,83.8)
    (2,84.2)
    (4,84.6)
    (7,85.3)
    (11,86.2)
    (16,86.8)
    };
    % RM 2
    \addlegendentry{\scalebox{.66}{GRAM-R$^2$ (LLaMA-3.2-3B-Ins)}}

    \end{axis}} 
    % \node [anchor=center] at (0.11\textwidth,-21.8em) {\scalebox{1.2}{(a) Adaptation (LLaMA-3.1-8B-Instruct)}};

  % fig2
  \scriptsize{
    \begin{axis}[
    at={(8em,-18em)},
    anchor=south west,
    ymajorgrids,
    xmajorgrids,
    grid style=dashed,
    width=.27\textwidth,
    height=.24\textwidth,
    xlabel={Sampling Size},
    ylabel={JudgeBench},
    ylabel style={scale=1.1,yshift=-2.0em},
    xlabel style={scale=1.1,yshift=0.75em},
    yticklabel style={/pgf/number format/fixed,/pgf/number format/fixed zerofill,/pgf/number format/precision=0,rotate=0,scale=1.0,anchor=east,align=right},
    ymin=79.8,
    ymax=84.2, 
    ytick={80,81,...,85},
    xmin=-0.5,
    xmax=16.5,
    xtick={0,2,4,7,11,16},
    xticklabels={1,8,16,32,64,128},  
    % 设置不等间隔的标签
    scaled ticks=false,
    x tick label style={
         anchor=center,
         scale=1.2,
         yshift=-0.8em
     },    
    y tick label style={
         anchor=east,
         scale=1.1,
         xshift=-.1em,
     }, 
    legend style={yshift=20pt,xshift=47.5em,font={\small},cells={anchor=west},fill opacity=0.8,legend columns=-1}
    ]

    \addplot[green, mark=triangle*,mark size=1.1pt, line width=1.4pt] coordinates {
    (0,81.0)
    (2,81.3)
    (4,81.6)
    (7,82.3)
    (11,83.5)
    (16,83.8)
    };
    % RM 1

    \addplot[green!50, mark=square*,mark size=1.1pt,line width=1.4pt] coordinates {
    (0,80.3)
    (2,80.5)
    (4,80.8)
    (7,81.7)
    (11,82.5)
    (16,82.8)
    };
    % RM 2

    \end{axis}} 

  % fig3
\end{tikzpicture}
    \vspace{-6mm}
    \caption{Test-time scaling performance of GRAM-R$^2$ using BoN sampling.}
    \label{fig:tts_gram_rr}
\end{figure}
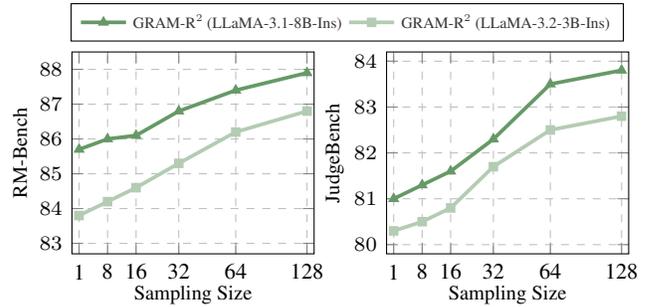

\begin{table}[!t]
    \centering
    \vspace{2mm}
    \resizebox{0.99\linewidth}{!}{% Please add the following required packages to your document preamble:
\begin{tabular}{lcccc}
\toprule[1.1pt]
\multirow{2}{*}{Method} & \multicolumn{1}{c}{\multirow{2}{*}{\begin{tabular}[c]{@{}c@{}}Reward\\ Reasoning\end{tabular}}} & \multicolumn{3}{c}{Data Used} \\ \cmidrule(l){3-5}
 & \multicolumn{1}{c}{} & RBD &RFD  & UD \\  \midrule
\multicolumn{5}{l}{\textit{\textbf{Discriminative Reward Models}}}  \\ 
POLAR \cite{dou2025pre} &  &   &\checkmark   &\checkmark   \\
WorldPM \cite{wang2025worldpm} &  &   &\checkmark   &   \\
GRM \cite{yang2024regularizing} & &  & \checkmark  \\
Skywork-Reward-v1 \cite{liu2024skywork} &                      &   &\checkmark   &    \\ 
Skywork-Reward-v2 \cite{liu2025skywork} &                      &   &\checkmark   &\checkmark    \\ \midrule
\multicolumn{5}{l}{\textit{\textbf{Generative Reward Models}}}  \\  
GRAM \cite{wang2025gram} &   &   &\checkmark   &\checkmark   \\
RM-R1 \cite{chen2025rm}  &\checkmark   &\checkmark   &\checkmark   &   \\
RRM \cite{guo2025reward} &\checkmark   &\checkmark   &  \checkmark &   \\
SyncPL \cite{liang2025generative} &\checkmark & \checkmark &  \checkmark & \\
Nemotron-Super \cite{wang2025helpsteer3} &\checkmark   &\checkmark   &\checkmark   &   \\
GRAM-R$^2$ &\checkmark  & \checkmark   &\checkmark   &\checkmark    \\ 
\bottomrule[1.1pt]
\end{tabular}}
    \caption{
    Existing reward model training approaches. Note that the use of RBD indicates whether the model is capable of leveraging annotated rationales during training.
    RBD: Rationale-based Labeled Data; RFD: Rationale-free Labeled Data; UD: Unlabeled Data.
    }
    \label{tab:exist_approaches_comparsion}
\end{table}

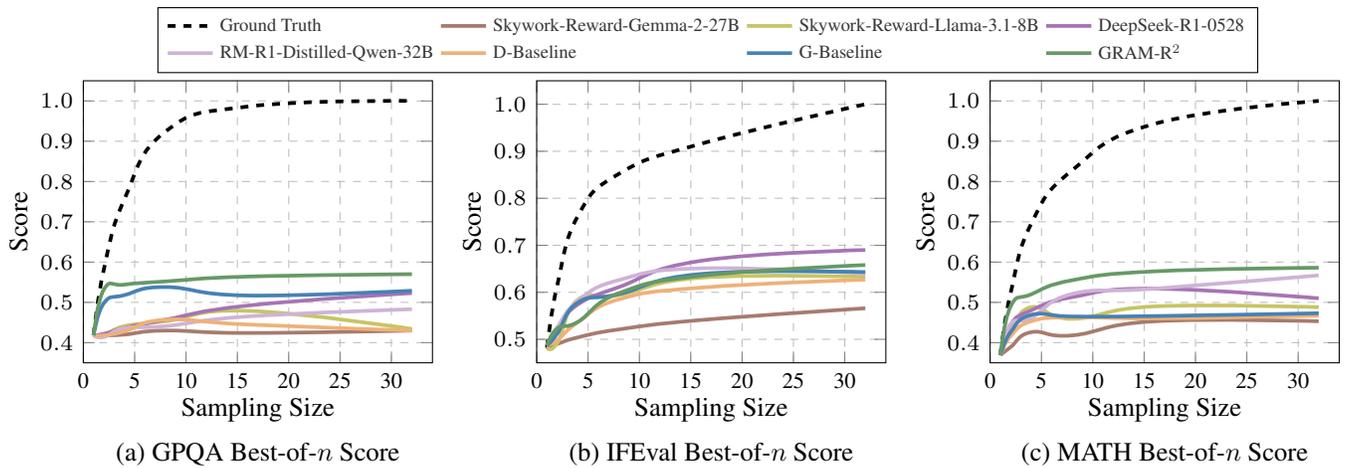
\begin{figure*}[!t]
    \centering
    % \definecolor{green}{RGB}{182,215,168}
% \definecolor{purple}{RGB}{157,193,230}
% \definecolor{blue}{RGB}{135,206,250}
% \definecolor{red}{RGB}{240,128,128}
\definecolor{green}{RGB}{102,153,102}  % green
\definecolor{purple}{RGB}{167,115,181}   % Darker version of the original purple
\definecolor{blue}{RGB}{70,130,180}   % blue
\definecolor{red}{RGB}{239,177,121}    % red
\definecolor{yellow}{RGB}{180,68,62}
\definecolor{brawn}{RGB}{166,123,110}
\definecolor{red2}{RGB}{207,179,215}
\definecolor{oran}{RGB}{199,197,99}

\begin{tikzpicture}
  \hspace{-1.0mm}
  % fig5
  \scriptsize{
    \begin{axis}[
    at={(-6em,-30em)},
    anchor=south west,
    ymajorgrids,
    xmajorgrids,
    grid style=dashed,
    width=.35\textwidth,
    height=.30\textwidth,
    xlabel={\normalsize Sampling Size},
    ylabel={\normalsize Score},
    ylabel style={yshift=-1.5em},
    xlabel style={yshift=0.6em},
    yticklabel style={/pgf/number format/fixed,/pgf/number format/fixed zerofill,/pgf/number format/precision=1,rotate=0,scale=1.0},
    ymin=0.35,
    ymax=1.05, 
    ytick={0.4,0.5,...,1.0},
    xmin=0,
    xmax=34,
    xtick={0,5,...,30},
    xticklabels={0,5,10,15,20,25,30},  % 设置不等间隔的标签
    scaled ticks=false,
    x tick label style={
         anchor=center,
         scale=1.2,
         yshift=-0.8em
     },    
    y tick label style={
         anchor=east,
         scale=1.2,
         xshift=-.1em,
     }, 
    legend style={yshift=30pt,xshift=45em,font={\normalsize},fill opacity=0.8,cells={anchor=west}, legend columns=4},
    title={(a) GPQA Best-of-$n$ Score},
    title style={yshift=-22em, font=\normalsize},
    ]
    \addplot[smooth, tension=0.8, black, dashed, mark=none, line width=1.4pt] coordinates {
    (1, 0.418)
    (2, 0.582)
    (4, 0.756)
    (8, 0.924)
    (16, 0.986)
    (32, 1.00)
    };
    % RM 1
    \addlegendentry{\scalebox{.7}{Ground Truth}};
    
    \addplot[smooth, tension=0.8, brawn, mark=none, line width=1.4pt] coordinates {
    (1, 0.418)
    (2, 0.418)
    (4, 0.420)
    (8, 0.430)
    (16, 0.424)
    (32, 0.430)
    };
    % RM 2
    \addlegendentry{\scalebox{.7}{Skywork-Reward-Gemma-2-27B}}

    \addplot[smooth, tension=0.8, oran, mark=none, line width=1.4pt] coordinates {
    (1, 0.418)
    (2, 0.418)
    (4, 0.441)
    (8, 0.453)
    (16, 0.479)
    (32, 0.434)
    };
    % RM 3
    \addlegendentry{\scalebox{.7}{Skywork-Reward-Llama-3.1-8B}}

    \addplot[smooth, tension=0.8, purple, mark=none, line width=1.4pt] coordinates {
    (1, 0.418)
    (2, 0.422)
    (4, 0.436)
    (8, 0.457)
    (16, 0.492)
    (32, 0.523)
    };
    % RM 4
    \addlegendentry{\scalebox{.7}{DeepSeek-R1-0528}}

    \addplot[smooth, tension=0.8, red2, mark=none, line width=1.4pt] coordinates {
    (1, 0.418)
    (2, 0.414)
    (4, 0.436)
    (8, 0.441)
    (16, 0.465)
    (32, 0.483)
    };
    % RM 5
    \addlegendentry{\scalebox{.7}{RM-R1-Distilled-Qwen-32B}}

    \addplot[smooth, tension=0.8, red, mark=none, line width=1.4pt] coordinates {
    (1, 0.418)
    (2, 0.418)
    (4, 0.428)
    (8, 0.457)
    (16, 0.445)
    (32, 0.430)
    };
    % RM 6
    \addlegendentry{\scalebox{.7}{D-Baseline}}

    \addplot[smooth, tension=0.8, blue, mark=none, line width=1.4pt] coordinates {
    (1, 0.418)
    (2, 0.498)
    (4, 0.518)
    (8, 0.538)
    (16, 0.517)
    (32, 0.529)
    };
    % RM 7
    \addlegendentry{\scalebox{.7}{G-Baseline}}

    \addplot[smooth, tension=0.8, green, mark=none, line width=1.4pt] coordinates {
    (1, 0.418)
    (2, 0.532)
    (4, 0.544)
    (8, 0.552)
    (16, 0.564)
    (32, 0.570)
    };
    % RM 8
    \addlegendentry{\scalebox{.7}{GRAM-R$^2$}}

    \end{axis}}

    % fig6
    \scriptsize{
    \begin{axis}[
    at={(18.5em,-30em)},
    anchor=south west,
    ymajorgrids,
    xmajorgrids,
    grid style=dashed,
    width=.35\textwidth,
    height=.30\textwidth,
    xlabel={\normalsize{Sampling Size}},
    ylabel={\normalsize{Score}},
    ylabel style={yshift=-1.5em},
    xlabel style={yshift=0.6em},
    yticklabel style={/pgf/number format/fixed,/pgf/number format/fixed zerofill,/pgf/number format/precision=1,rotate=0,scale=1.0},
    ymin=0.45,
    ymax=1.05, 
    ytick={0.5,0.6,...,1.0},
    xmin=0,
    xmax=34,
    xtick={0,5,...,30},
    xticklabels={0,5,10,15,20,25,30},  % 设置不等间隔的标签
    scaled ticks=false,
    clip=false,
    x tick label style={
         anchor=center,
         scale=1.2,
         yshift=-0.8em
     },    
    y tick label style={
         anchor=east,
         scale=1.2,
         xshift=-.1em,
     }, 
    title={(b) IFEval Best-of-$n$ Score},
    title style={yshift=-22em, font=\normalsize},
    ]

    \addplot[smooth, tension=0.8, black, dashed, mark=none, line width=1.4pt] coordinates {
    (1, 0.482)
    (2, 0.615)
    (4, 0.762)
    (8, 0.852)
    (16, 0.916)
    (32, 1.000)
    };
    % RM 1
    \addplot[smooth, tension=0.8, brawn, mark=none, line width=1.4pt] coordinates {
    (1, 0.482)
    (2, 0.490)
    (4, 0.504)
    (8, 0.521)
    (16, 0.541)
    (32, 0.566)
    };
    % RM 2
    \addplot[smooth, tension=0.8, oran, mark=none, line width=1.4pt] coordinates {
    (1, 0.482)
    (2, 0.492)
    (4, 0.584)
    (8, 0.594)
    (16, 0.631)
    (32, 0.633)
    };
    % RM 3
    \addplot[smooth, tension=0.8, purple, mark=none, line width=1.4pt] coordinates {
    (1, 0.487)
    (2, 0.524)
    (4, 0.574)
    (8, 0.611)
    (16, 0.667)
    (32, 0.690)
    };
    % RM 4
    \addplot[smooth, tension=0.8, red, mark=none, line width=1.4pt] coordinates {
    (1, 0.487)
    (2, 0.498)
    (4, 0.536)
    (8, 0.585)
    (16, 0.610)
    (32, 0.627)
    };
    \addplot[smooth, tension=0.8, red2, mark=none, line width=1.4pt] coordinates {
    (1, 0.487)
    (2, 0.525)
    (4, 0.581)
    (8, 0.626)
    (16, 0.651)
    (32, 0.640)
    };
    % RM 5
    % RM 6
    \addplot[smooth, tension=0.8, blue, mark=none, line width=1.4pt] coordinates {
    (1, 0.487)
    (2, 0.512)
    (4, 0.578)
    (8, 0.596)
    (16, 0.639)
    (32, 0.643)
    };
    % RM 7
    \addplot[smooth, tension=0.8, green, mark=none, line width=1.4pt] coordinates {
    (1, 0.487)
    (2, 0.523)
    (4, 0.537)
    (8, 0.598)
    (16, 0.635)
    (32, 0.658)
    };
    % RM 8
    
    \end{axis}  
   }
    % \node [anchor=center] at (.0\textwidth+18.em,-21.6em) {\scalebox{1.0}{(f) RL (LLaMA-3.1-8B-Instruct)}};
  % fig7
  \scriptsize{
    \begin{axis}[
   at={(43em,-30em)},
   anchor=south west,
    ymajorgrids,
    xmajorgrids,
    grid style=dashed,
    width=.35\textwidth,
    height=.30\textwidth,
    xlabel={\normalsize{Sampling Size}},
    ylabel={\normalsize{Score}},
    ylabel style={yshift=-1.5em},
    xlabel style={yshift=0.6em},
    yticklabel style={/pgf/number format/fixed,/pgf/number format/fixed zerofill,/pgf/number format/precision=1,rotate=0,scale=1.0},
    ymin=0.35,
    ymax=1.05, 
    ytick={0.3,0.4,...,1.0},
    xmin=0,
    xmax=34,
    xtick={0,5,...,30},
    xticklabels={0,5,10,15,20,25,30},  % 设置不等间隔的标签
    x tick label style={
         anchor=center,
         scale=1.2,
         yshift=-0.8em
     },    
    y tick label style={
         anchor=east,
         scale=1.2,
         xshift=-.1em,
     }, 
    title={(c) MATH Best-of-$n$ Score},
    title style={yshift=-22em, font=\normalsize},
    ]
    \addplot[smooth, tension=0.8, black, dashed, mark=none, line width=1.4pt] coordinates {
    (1, 0.369)
    (2, 0.523)
    (4, 0.695)
    (8, 0.828)
    (16, 0.943)
    (32, 1.00)
    };
    % RM 1
    \addplot[smooth, tension=0.8, brawn, mark=none, line width=1.4pt] coordinates {
    (1, 0.369)
    (2, 0.389)
    (4, 0.426)
    (8, 0.418)
    (16, 0.453)
    (32, 0.453)
    };
    % RM 2
    \addplot[smooth, tension=0.8, oran, mark=none, line width=1.4pt] coordinates {
    (1, 0.369)
    (2, 0.426)
    (4, 0.488)
    (8, 0.459)
    (16, 0.490)
    (32, 0.488)
    };
    % RM 3
    \addplot[smooth, tension=0.8, purple, mark=none, line width=1.4pt] coordinates {
    (1, 0.369)
    (2, 0.445)
    (4, 0.480)
    (8, 0.513)
    (16, 0.534)
    (32, 0.510)
    };
    % RM 4
    \addplot[smooth, tension=0.8, red2, mark=none, line width=1.4pt] coordinates {
    (1, 0.369)
    (2, 0.446)
    (4, 0.461)
    (8, 0.521)
    (16, 0.534)
    (32, 0.567)
    };
    % RM 5
    \addplot[smooth, tension=0.8, red, mark=none, line width=1.4pt] coordinates {
    (1, 0.369)
    (2, 0.416)
    (4, 0.453)
    (8, 0.463)
    (16, 0.460)
    (32, 0.467)
    };
    % RM 6
    \addplot[smooth, tension=0.8, blue, mark=none, line width=1.4pt] coordinates {
    (1, 0.369)
    (2, 0.426)
    (4, 0.469)
    (8, 0.465)
    (16, 0.466)
    (32, 0.473)
    };
    % RM 7
    \addplot[smooth, tension=0.8, green, mark=none, line width=1.4pt] coordinates {
    (1, 0.369)
    (2, 0.490)
    (4, 0.520)
    (8, 0.555)
    (16, 0.577)
    (32, 0.586)
    };
    % RM 8

    \end{axis}
   }

\end{tikzpicture}
    \vspace{-5mm}
    \caption{
    The performance of reward models on the PPE benchmark. For the results, we can observe that our GRAM-R$^2$ can achieve superior performance when used to select the best response from multiple candidates on several challenging downstream benchmarks, including GPQA, IFEval, and MATH. This demonstrates its significant potential to enhance the problem-solving and reasoning capabilities of LLMs through test-time re-ranking.
    }
    \label{fig:res-list-wise-ranking-appendix}
\end{figure*}

\subsection{Comparison of Existing Reward Model Training Approaches}
We conduct a comparative analysis of recent reward model training approaches across several key dimensions: their inclusion of reward reasoning, their capacity to leverage unlabeled data, their utilization of rationale-based labeled data, and their use of rationale-free labeled data. This comparison is summarized in Table~\ref{tab:exist_approaches_comparsion}. As the table illustrates, our reward model training approach is unique in its ability to holistically integrate these diverse data types into a single, cohesive training approach. This versatility highlights the significant potential of our approach in maximizing data utility, demonstrating a clear advantage in how preference data of various formats can be leveraged to construct more powerful and robust reward models.

% \begin{table}[!t]
%     \centering
%     \caption{111}
%     \label{tab:comparsion_with_skywork_v2}
% \end{table

% \begin{table*}[!t]
%     \centering
%     \resizebox{0.98\linewidth}{!}{
%     \input{tables/res_gram_rr_with_other_backbones}}
%     \caption{
%     }
% \end{table*}

\clearpage

\begin{figure*}[!t]
    \centering
    \input{images/merge_feedback_template}
    \caption{
    Template used for merging the feedback.
    }
    \label{fig:merge-feedback}
\end{figure*}

\clearpage

\begin{figure*}[!t]
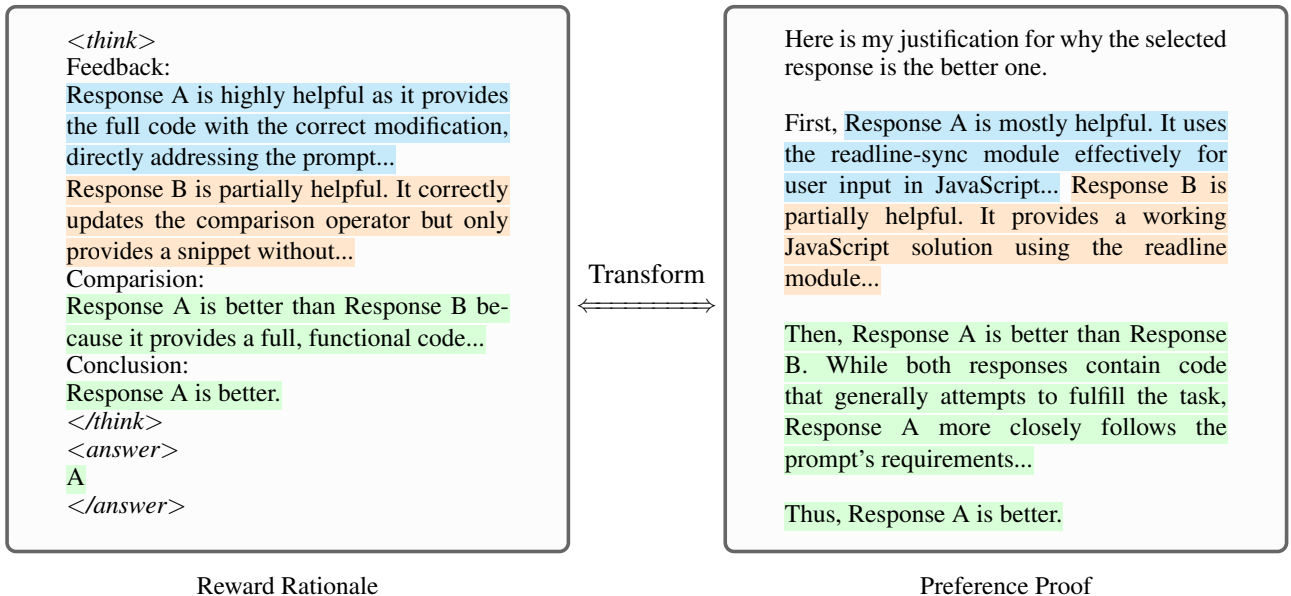

    \centering
    \begin{subfigure}[b]{\textwidth} 
        \centering
        \scalebox{0.96}{\input{images/convert_template}} 
        \subcaption{Template used for training the preference-proving model.}
        \label{fig:sub_a}
    \end{subfigure}
    
    \vspace{5pt} 
    
    \begin{subfigure}[b]{\textwidth}  
        \centering
        \scalebox{0.96}{\input{images/template_preference_proving_model}} % 
        \subcaption{A complete example demonstrating the conversion between a rationale and a proof.}
        \label{fig:sub_b}
    \end{subfigure}
    \caption{
   Sub-figure (a) illustrates the template used for training the preference-proving model. Sub-figure (b) presents a complete example demonstrating the conversion between a rationale and a proof.
    }
    \label{fig:template-preference-proving-model}
\end{figure*}

\begin{figure*}[!t]
    \centering
    \input{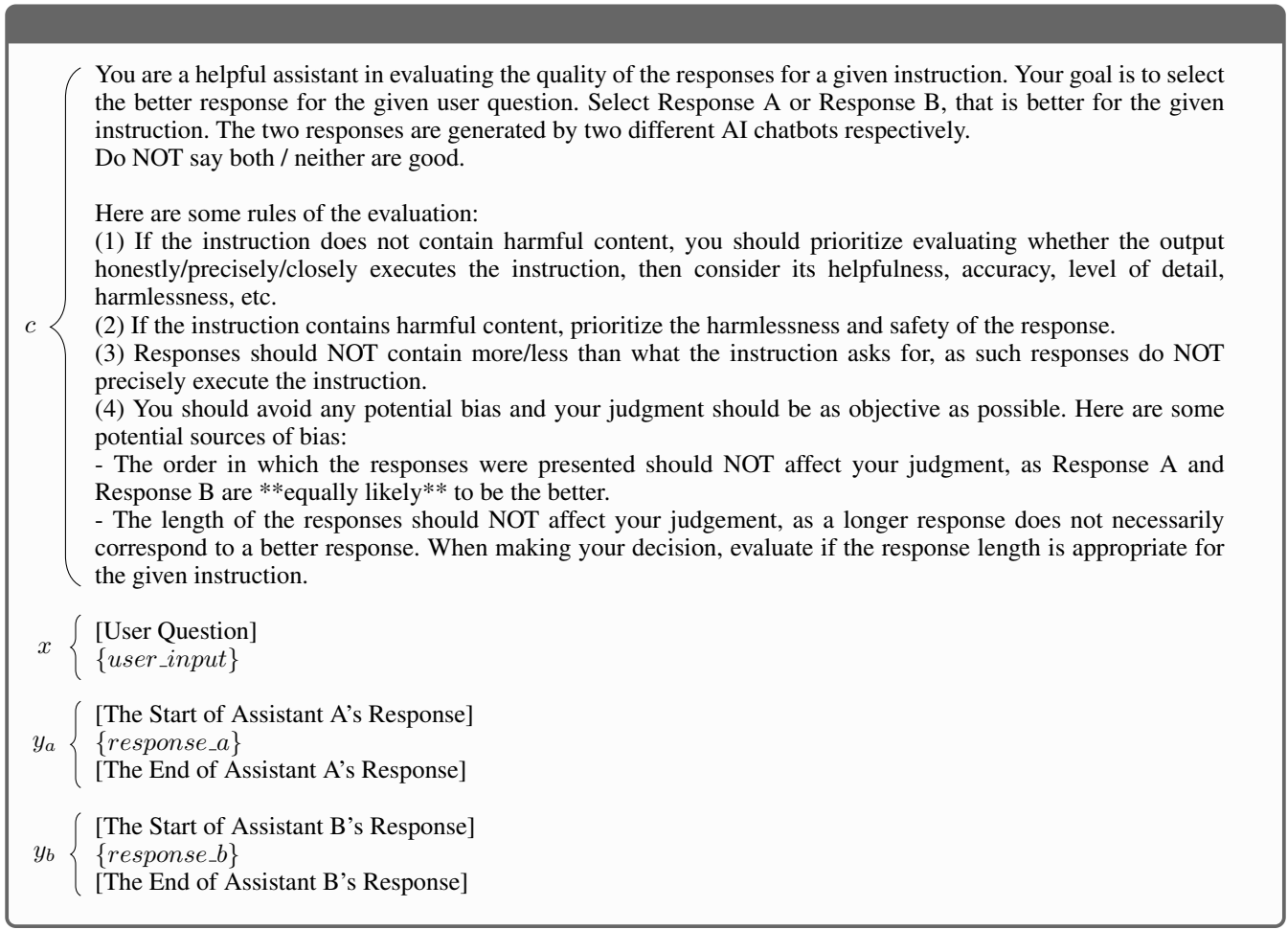}
    \caption{
    Template used for training the generative baseline reward model. Note that the prompt format is kept consistent with that used in GRAM-R$^2$ training to ensure a fair comparison.
    }
    \label{fig:template-g-baseline}
\end{figure*}

\begin{figure*}[t]
    \centering
    \input{images/template_gram_rr}
    \caption{
    Template used for training our GRAM-R$^2$.
    }
    \label{fig:template-our-gram-rr}
\end{figure*}

\clearpage

\begin{table*}[!t]
    \centering
    \resizebox{0.95\linewidth}{!}{
    \begin{tabular}{ll}
\toprule[1.1pt]
\bf{Input} &  
\parbox{0.75\linewidth}{
What's the permission that allows creating provisioning profiles in Apple Developer account is called?
} \\ \midrule
SFT   &
\parbox{0.75\linewidth}{
The permission is called "Access to Certificates, Identifiers \& Profiles" in the Apple Developer account.
}     \\  \midrule
D-Baseline    & 
\parbox{0.75\linewidth}{
To create provisioning profiles in an Apple Developer account, a user needs the permission called "Access to Certificates, Identifiers \& Profiles." This pession allows users to manage certificates, such as App IDs. This pession allows users to manage certificates, such as App IDs.
}      \\ \midrule
G-Baseline          &
\parbox{0.75\linewidth}{
The permission is called "Access to Certificates, Identifiers \& Profiles".
}
     \\  \midrule
GRAM-Qwen3-8            & 
\parbox{0.75\linewidth}{
The permission required to create provisioning profiles in an Apple Developer account is called "Access to Certificates, Identifiers \& Profiles", which grants the user the ability to manage certificates, App IDs, and provisioning profiles.
}
\\  \midrule
Skywork-Reward-8B   &
\parbox{0.75\linewidth}{
The permission required to create provisioning profiles in an Apple Developer account is called "Access to Certificates, Identifiers Identifiers Identifiers.....
}
\\  \midrule
GRAM-R$^2$                &  
\parbox{0.75\linewidth}{
The permission that allows creating provisioning profiles in an Apple Developer account is called "Access to Certificates, Identifiers \& Profiles". This permission allows the user to manage provisioning profiles, certificates, and App IDs. Without this permission, the user cannot create or edit provisioning profiles in the developer portal.
}
\\ \bottomrule[1.1pt]
\end{tabular}}
    \caption{
    A case in the PPO fine-tuning experiments. Compared to baseline reward models, which lead to incomplete, inaccurate, or repetitive responses, the model fine-tuned with GRAM-R$^2$ produces a more accurate and comprehensive answer, demonstrating better alignment with human intent.
    }
\end{table*}

\begin{table*}[!t]
    \centering
    \resizebox{0.95\linewidth}{!}{
    \begin{tabular}{ll}
\toprule[1.1pt]
\bf{Input} &  
\parbox{0.75\linewidth}{
there used to be a program for winmx called mcma, what does mcma stand for and what does the program do?
} \\ \midrule
SFT   &
\parbox{0.75\linewidth}{
MCMA stands for "Music Content Management Assistant" and it was a tool that allowed users to illegally download paid software directly from WinMX's servers without any restrictions.
}     \\  \midrule
D-Baseline    & 
\parbox{0.75\linewidth}{
MCMA, short for "Media Content Monitoring App", was developed in 2015 as a plugin for WinMX, mainly used to track users' download histories and secretly share their private files with third-party servers. 
}      \\ \midrule
G-Baseline          &
\parbox{0.75\linewidth}{
MCMA stands for Motion Control \& Motor Association, which was founded in 2008 by the merger of the Motion Control Association and the Motor \& Motion Association. It's a tool for WinMX that helps users manage their music files and it can also be used to download movies directly from WinMX in 2020 without any limitations.
}
     \\  \midrule
GRAM-Qwen3-8            & 
\parbox{0.75\linewidth}{
MCMA stands for "Multi Connection Manager Application." It was a handy tool that many WinMX users found useful back in the day. Basically, it helped manage multiple simultaneous connections to improve file sharing speeds and reliability.
}
\\  \midrule
Skywork-Reward-8B            & 
\parbox{0.75\linewidth}{
MCMA stands for "WinMX Community Management Application," and it was a program used to manage and enhance the functionality of the WinMX peer-to-peer file-sharing network.
}
\\  \midrule
GRAM-R$^2$   &
\parbox{0.75\linewidth}{
MCMA was an external utility created for WinMX, the once-popular P2P file-sharing client of the early 2000s. It enabled users to launch several WinMX instances on a single machine, each tied to a distinct account. By signing multiple accounts into the WinMX network simultaneously, users could boost their overall bandwidth or sharing potential through the extra concurrent connections.
}
\\ \bottomrule[1.1pt]
\end{tabular}}
    \caption{
    A case in the PPO fine-tuning experiments. The model fine-tuned with GRAM-R$^2$ delivers the most functionally accurate and informative description of the obscure software (WinMX MCMA). In contrast, the baseline models exhibit severe factual errors and hallucinations. For instance, G-Baseline confuses the software with an entirely unrelated real-world association, while D-Baseline fabricates a malicious purpose.
    }
\end{table*}

\begin{table*}[!t]
    \centering
    \begin{tabular}{ll}
\toprule[1.1pt]
\bf{Input} &  
\parbox{0.75\linewidth}{
A college student initiated a criminal case against his former roommate, claiming the former roommate snuck in and removed his state-of-the-art home theater equipment from his house. ... ... Once you have your answer, please duplicate that letter five times in a single string. For example, if the answer is K, then write KKKKK. Let's think step by step.
} \\ \midrule
\textbf{Response A}   &
\parbox{0.75\linewidth}{
To determine if evidence of the student's reputation for dishonesty is admissible in a claim of malicious prosecution, we need to consider several legal principles,  ... ... The best-balanced reasoning aligns with choice (F), as it clearly states that when character is directly at issue, such as in a malicious prosecution claim, it becomes admissible. Hence, my answer is: FFFFF
}     \\  \midrule
\textbf{Response B}    & 
\parbox{0.75\linewidth}{
To determine whether the former roommate can use evidence of the college student's reputation for dishonesty in his civil case for malicious prosecution, we need to consider rules of admissibility of character evidence, ... ... Based on these considerations, the best answer is:(A) Yes, because the student's reputation for dishonesty directly relates to the tort of malicious prosecution. So, the correct and best-supported answer is: AAAAA
}      \\ \midrule
\textbf{Rationale}         &
\parbox{0.75\linewidth}{
\textit{$<$think$>$}Feedback:

Response A is perfectly helpful. It provides a clear, comprehensive, and accurate analysis of the legal principles involved in character evidence admissibility for malicious prosecution, correctly selecting (F) as the answer. The response is well-structured, uses intermediate language, and is moderately lengthy, ensuring accessibility and clarity. It also aligns with the prompt's request for a multiple-choice answer and includes a step-by-step analysis of each option. Response B is partially helpful. It offers a detailed analysis of character evidence in malicious prosecution and correctly identifies the importance of dishonesty in the case. However, it incorrectly selects (A) as the answer, missing the nuance of character evidence admissibility. The response is well-written but could be more concise and accurate in its legal reasoning.

Comparison: 

Response A is better than Response B. It provides a more accurate and nuanced analysis of character evidence admissibility in malicious prosecution, correctly identifying (F) as the answer, while Response B incorrectly selects (A). Response A also offers a clearer, more structured explanation of the legal principles involved, aligning better with the prompt's requirements.

Conclusion:

Response A is better.
\textit{$<$/think$>$}

\textit{$<$answer$>$}A\textit{$<$/answer$>$}
}
\\ \bottomrule[1.1pt]
\end{tabular}
    \caption{
    Case one of a rationale generated by GRAM-R$^2$. The ground truth preference label is `A' because response A provides the correct answer and is clearly articulated. As shown, the GRAM-R$^2$ model can successfully capture these qualities and predict the correct preference label.
    }
\end{table*}

\begin{table*}[!t]
    \centering
    \begin{tabular}{ll}
\toprule[1.1pt]
\bf{Input} &  
\parbox{0.75\linewidth}{
Compute the geometric mean of $\{8, -10\}$. Please put your final answer in a $boxed{}$.
} \\ \midrule
\textbf{Response A}   &
\parbox{0.75\linewidth}{
The geometric mean of a set of \( n \) numbers \( a_1, a_2, \ldots, a_n \) is given by:\[\sqrt[n]{a_1 \cdot a_2 \cdot \cdots \cdot a_n}\] To find the geometric mean of the numbers \( \{8, -10\} \) :\[\sqrt[2]{8 \cdot (-10)}\] First, compute the product inside the square root :\[8 \cdot (-10) = -80\] Next, compute the square root of \( -80 \). The square root of a negative number involves complex numbers. Specifically :\[\sqrt{-80} = \sqrt{-1 \cdot 80} = \sqrt{-1} \cdot \sqrt{80} = i \cdot \sqrt{80}\] Simplify \(\sqrt{80}\) :\[\sqrt{80} = \sqrt{16 \times 5} = \sqrt{16} \times \sqrt{5} = 4\sqrt{5}\] Thus:\[\sqrt{-80} = i \cdot 4\sqrt{5} = 4i\sqrt{5}\] The geometric mean of \( \{8, -10\} \) is:\[\boxed{4i\sqrt{5}}\]
}     \\  \midrule
\textbf{Response B}    & 
\parbox{0.75\linewidth}{
The geometric mean of a set of numbers is found by multiplying the numbers together and then taking the \(n\)-th root of the product, where \(n\) is the number of values in the set. Let's determine the geometric mean of the set \(\{8, -10\}\). First, we calculate the product of the numbers:\[8 \times (-10) = -80\]Next, we need to take the square root of \(-80\), because there are 2 numbers in the set. However, the square root of a negative number is not defined in the set of real numbers. Therefore, the geometric mean of a set that includes negative numbers is not defined when considering only real numbers. In summary, the geometric mean of the set \(\{8, -10\}\) is not defined in the context of real numbers. Thus, the final answer is:\[\boxed{\text{undefined}}\]
}      \\ \midrule
\textbf{Rationale}         &
\parbox{0.75\linewidth}{
$<$think$>$Feedback:Response A is perfectly helpful. It correctly defines the geometric mean and applies the correct formula, including the use of complex numbers for negative inputs. The explanation is detailed and accurate, though the formatting could be improved for readability. The response fully addresses the prompt and provides a correct solution. Response B is not helpful. It incorrectly states that the geometric mean is undefined for negative numbers, which is false. The geometric mean can be calculated using complex numbers, and the response fails to provide this solution. Additionally, the formatting is unclear, making the response difficult to follow. Comparison: Response A is much better than Response B. Response A provides a correct and detailed explanation of how to calculate the geometric mean, including handling negative numbers with complex numbers. Response B incorrectly claims the geometric mean is undefined for negative numbers, which is factually wrong. Additionally, Response B fails to provide the correct solution using complex numbers, while Response A does so accurately. Conclusion:Response A is better.$<$/think$>$$<$answer$>$A$<$/answer$>$
}
\\ \bottomrule[1.1pt]
\end{tabular}
    \caption{
    Case one of a rationale generated by GRAM-R$^2$. The ground-truth preference label is `A' because Response A follows the instruction ``Please put your final answer in a \textit{boxed}'', and additionally provides both the correct calculation steps and the final answer. As shown, the GRAM-R$^2$ model can effectively capture these differences between Response A and Response B, and consequently generate an accurate preference.
    }
\end{table*}

% 讨论偏好分歧的问题

\end{document}